\documentclass{article}

\usepackage{arxiv}
\usepackage[square,numbers]{natbib}
\usepackage[utf8]{inputenc} 
\usepackage[T1]{fontenc}    
\usepackage{hyperref}       
\usepackage{url}            
\usepackage{booktabs}       
\usepackage{amsfonts}       
\usepackage{nicefrac}       
\usepackage{microtype}      
\usepackage{lipsum}		
\usepackage{graphicx}
\usepackage{doi}
\usepackage{tabularx}
\usepackage{graphicx}
\usepackage{subcaption}
\usepackage{amsmath}
\usepackage{mdframed}
\usepackage{longtable}
\newcolumntype{P}[1]{>{\raggedright\arraybackslash}p{#1}}
\usepackage{comment}
\usepackage{csquotes}

\usepackage{multirow}
\usepackage{array}
\usepackage{listings}
\usepackage{soul}
\usepackage{xcolor}

\usepackage{fancyvrb}

\title{Automating Research Synthesis with Domain-Specific Large Language Model Fine-Tuning}

\author{ \href{https://orcid.org/0000-0001-9416-1435}{\includegraphics[scale=0.06]{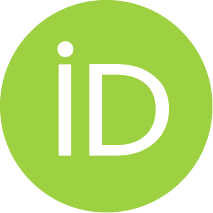}\hspace{1mm}Teo ~Susnjak} \thanks{Corresponding author: t.susnjak@massey.ac.nz} \\
	School of Mathematical and Computational Sciences\\
	Massey University\\
	Albany, New Zealand \\
	\And
	Peter Hwang \\
	School of Mathematical and Computational Sciences\\
	Massey University\\
	Albany, New Zealand \\
	\And
	\href{https://orcid.org/0000-0002-0683-436X}{\includegraphics[scale=0.06]{orcid.pdf}\hspace{1mm}Napoleon H. Reyes}\\
	School of Mathematical and Computational Sciences\\
	Massey University\\
	Albany, New Zealand \\
\And
	\href{https://orcid.org/0000-0001-7648-285X}{\includegraphics[scale=0.06]{orcid.pdf}\hspace{1mm}Andre L. C. Barczak}\\
	  Centre for Data Analytics\\
	Bond University\\
	Gold Coast, Australia \\
	\And
 \And \And \And \And \And\And
	\href{https://orcid.org/0000-0003-0836-4266}{\includegraphics[scale=0.06]{orcid.pdf}\hspace{1mm}Timothy R. McIntosh} \\
	Cyberoo Pty Ltd \\
	 Surry Hills, NSW\\
    Australia \\
  	\And
	\href{https://orcid.org/0000-0003-0701-0204}{\includegraphics[scale=0.06]{orcid.pdf}\hspace{1mm}Surangika Ranathunga   }\\
	School of Mathematical and Computational Sciences\\
	Massey University\\
	Albany, New Zealand \\
}

\renewcommand{\shorttitle}{Automating Research Synthesis with Domain-Specific LLM Fine-Tuning}

\begin{document}
\maketitle

\begin{abstract}
This research pioneers the use of finetuned Large Language Models (LLMs) to automate Systematic Literature Reviews (SLRs), presenting a significant and novel contribution in integrating AI to enhance academic research methodologies. Our study employed the latest finetuning methodologies together with open-sourced LLMs, and demonstrated a practical and efficient approach to automating the final execution stages of an SLR process that involves knowledge synthesis. The results maintained high fidelity in factual accuracy in LLM responses, and were validated through the replication of an existing PRISMA-conforming SLR. Our research proposed solutions for mitigating LLM hallucination and proposed mechanisms for tracking LLM responses to their sources of information, thus demonstrating how this approach can meet the rigorous demands of scholarly research. The findings ultimately confirmed  the potential of finetuned LLMs in streamlining various labour-intensive processes of conducting literature reviews. Given the potential of this approach and its applicability across all research domains, this foundational study also advocated for updating PRISMA reporting guidelines to incorporate AI-driven processes, ensuring methodological transparency and reliability in future SLRs. This study broadens the appeal of AI-enhanced tools across various academic and research fields, setting a new standard for conducting comprehensive and accurate literature reviews with more efficiency in the face of ever-increasing volumes of academic studies. 
\end{abstract}

\keywords{LLM Fine-tuning for SLRs  \and SLR Automation \and Retrieval-Augmented Generation for Research \and Domain-Specific Model Training \and Knowledge Synthesis AI  \and AI-Driven Research Synthesis \and Literature Review Automation \and Generative AI \and AI-Enhanced Systematic Reviews  \and PRISMA and AI Integration}

\section{Introduction}

Systematic Literature Reviews (SLRs) serve as the bedrock of academic research, playing a crucial role in the amalgamation, examination, and synthesis of existing scholarly knowledge across various fields\cite{paez2017Grey,Xiao2017Guidance,Siddaway2019How}. These reviews offer a methodical and replicable approach, ensuring the integrity and thoroughness of research synthesis especially when combined with reporting guidelines like PRISMA \cite{page2021prisma,Liberati2009The,Rethlefsen2021PRISMAS}. Such a foundation is indispensable for advancing both theoretical understanding and practical applications. However, the traditional execution of SLRs is marked by its manual and resource-intensive nature, often stretching over extensive periods, which introduces significant inefficiencies into the research process \cite{Williams2020Reexamining,de2023artificial}. 

The rigorous yet cumbersome character of traditional SLR methodologies presents considerable bottlenecks in the management and synthesis of large datasets of selected studies that hinges on effective information retrieval \cite{Chen2019Visualizing,Ganann2010Expediting}. These challenges not only prolong the synthesis - the execution phase of a review - but also hamper the ongoing updates of the SLRs with newer findings, and risk diminishing the timeliness and relevance of the insights gleaned\cite{McInnes1999Challenges,Mahood2014Searching}. This scenario underscores the need for innovative, scalable and sustainable solutions that can streamline the extraction of information from findings situated in academic papers, as well as its persistence in suitable information technologies which facilitate their accurate and effective retrieval necessary for executing SLRs \cite{Elamin2009Choice,bui2016extractive}.  

The recent advent of a new class of Artificial Intelligence (AI) systems like Large Language Models (LLMs), heralds a new epoch with the potential to dramatically redefine the SLR landscape through the automation of the information retrieval processes while maintaining high factual fidelity \cite{susnjak2023prisma,de2023artificial}. These models, with their advanced natural language comprehension capabilities, text generation and knowledge retention \cite{bonan2023recent}, offer a promising avenue for automating and optimizing various stages of the SLR process \cite{Yupeng2024llmsurvey}, and in particular the execution phase that relies on \enquote{talking to} both individual academic papers via LLMs, as well as simultaneously \enquote{talking across} all target papers for synthesising purposes \cite{Hou2023Large}. Despite their potential, the broad generalist pretraining of these models which have been trained on vast amounts of diverse text data means that the LLMs fall short in providing the domain-specific accuracy and precision in the information retrieved that is essential for the detailed task of knowledge synthesis across a very specific and narrow sets of studies. Additionally, their current propensity to hallucinate \cite{kalai2023calibrated,McIntosh2023hallucination,Li2023Evaluating,Nashwan2023Streamlining} renders them unable to consistently respond accurately, which precludes them from providing reliable responses needed to conduct SLRs with integrity. Additionally, current LLMs have variable abilities to audit and track the sources of their responses \cite{Khraisha2023Can}. Their inability to reliably ensure that the provenance of the LLM responses can be linked to the target studies that comprise an SLR corpus represents a serious limitation for using this technology for this purpose. Collectively, these gaps highlight a critical area within AI-assisted SLR processes for the purpose of enhancing the information retrieval capabilities of LLMs \cite{Fernandez2023How}. Indeed, resent research has repeatedly raised these concerns with respect to using LLM for the purposes of SLRs, especially in the knowledge synthesis stages. While it has been  suggested that LLMs could be used for assisting evidence synthesis tasks  in SLRs via their summarization capabilities, concerns have been raised about their lack of continuous learning capability and temporal reasoning \cite{peng2023ai,smith2024reviews}. \citet{qureshi2023chatgpt} noted that using LLMs shows promise for aiding in systematic review-related tasks, but the authors concluded that the technology is in its infancy and requires significant development for such applications. In their most recent review, \citet{bolanos2024artificial} found that LLM usage for SLRs is hampered by limitations such as reduced efficacy in domain-specific and narrower subject areas, their propensity for hallucinating and generating misleading information, alongside their opaque decision processes which cannot be audited. 

To surmount these challenges, this study proposes the creation of finetuned open-source LLMs, trained on the corpus of selected academic papers for a target SLR, expanding the generalist knowledge of an LLM with narrower domain-specific expertise. This work devises a novel way to automatically extract information from a set of academic papers in order to create SLR-specific datasets which can be leveraged for finetuning LLMs so that they can support question and answering downstream tasks. We also devise mechanisms to mitigate LLM hallucination and to ensure that all LLM responses related to an SLR can be tracked to source studies. 
The ensuing research presents a comprehensive SLR-automation framework with a focus on the knowledge synthesis stage that
aims to revolutionize information retrieval mechanisms with empirical evidence of their effectiveness, thereby expediting and transforming the synthesis of research findings in the context of SLRs.

\subsubsection*{Contribution}

Our study substantially contributes to the field of information retrieval in general where factual recall from arrays of documents is required and needs to be expressed in natural language, while possessing the ability to audit the responses with respect to the source information. Our proposed framework is specifically applied and demonstrated to the context of SLR-automation, where the goal was to validate the proposed framework through an empirical study that seeks to replicate a previously published PRISMA-conforming SLR, which serves as a gold standard and a test use case for validating the framework. The contributions of this research can be summarised as:

\begin{itemize}
\item Devising a methodical and automate approach for converting selected academic papers into datasets that can be used for finetuning LLMs. 
\item Proposing an effective approach to ensuring factual recall in responses by providing a mechanism to audit the source information of the LLM's response. 
\item Developed and validated evaluation metrics with the goal of testing for factuality in the LLM responses. 
\item Benchmarked various AI automation methodologies for SLRs, focusing on the comparative effectiveness of different finetuning approaches.
\item Demonstrated the efficacy of these proposed methodologies by replicating a published PRISMA-conforming SLR.
\item Enhancing the scholarly toolkit for SLRs with advanced, efficient, and context-aware AI technologies.
\item Setting new standards in academic research for reliability, validity, and ethical AI employment through pioneering AI methodologies.
\item Releasing a Python package \footnote{https://github.com/peterjhwang/slr-helper} to facilitate data curation for LLM fine-tuning, tailored for the unique requirements of SLRs.
\end{itemize}

\section{Background}

This literature review examines the SLR process, especially the knowledge synthesis phase and focuses on the integration of LLMs and AI to automate its various stages. It also addresses the current efforts in SLR automation, the role of LLMs in enhancing these processes, and the critical challenge of ensuring factual accuracy arising from LLM hallucinations. The review further explores studies that have considered the potential of fine-tuning domain-specific LLMs to tailor their performance for SLR tasks, aiming to provide a clear, concise overview of the advancements and challenges in employing AI tools for SLR automation.

\subsection{Systematic Literature Review Process and Synthesis of Studies}

\begin{figure}[hbt]
    \centering
    \includegraphics[width=.9\textwidth]{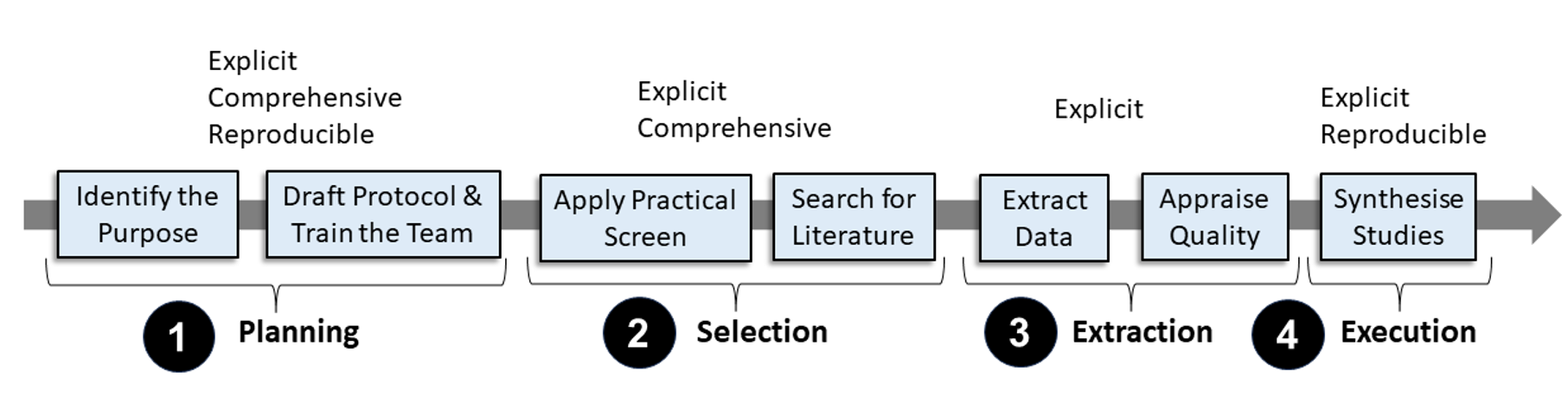}
    \caption{The general process of conducting an SLR as outlined by \citet{okoli2015guide}, denoting which steps should be explicitly reported to the reader, as well as explained to justify the comprehensiveness of the SLR despite exclusion criteria together with the detailed communication of steps taken to make the SLR reproducible.}
    \label{fig:diagramSLR}
\end{figure}

The process of conducting an SLR (Figure \ref{fig:diagramSLR}) begins with the clear articulation of the review's purpose and intended outcomes. This foundational step ensures that the review's scope and objectives are explicit to its audience \cite{okoli2015guide}. Following this, the development of a detailed protocol and the training of the review team are essential to guarantee methodological rigor and consistency in the review execution. The screening process, both for inclusion and quality appraisal, demands transparency in the criteria used for study selection and exclusion, ensuring that only the most relevant and high-quality studies are considered for synthesis. The comprehensive search for literature, coupled with systematic data extraction from the included studies, lays the groundwork for the subsequent synthesis phase. This phase is where the insights gleaned from individual studies are integrated to form a coherent narrative or to distill new theoretical understandings.

\subsubsection{Synthesis of Studies in SLRs}

Having navigated the initial steps of the SLR process, from purpose identification through to quality appraisal, we arrive at the execution part of the SLR, which is the synthesis phase (Step 4 in Figure \ref{fig:diagramSLR}). This juncture is where the accumulated evidence is methodically combined to reveals new insights or to forge theoretical advancements. The approach to synthesis is inherently dependent on the nature of the primary studies under consideration—quantitative, qualitative, or a combination thereof \cite{grant2009typology}. 
The synthesis phase in SLRs is arguably one of the most challenging yet crucial stages of the literature review process. It demands not only a comprehensive understanding of the included studies but also a demanding task of integrating their findings \cite{torraco2016writing}. This phase is pivotal for transcending beyond mere aggregation of data and information and instead aiming to distill new insights or theoretical contributions from the collective evidence \cite{torraco2005writing}. The complexity of this task is magnified by the diversity of study designs, methodologies, terminologies, and data types encountered within the corpus of literature being reviewed \cite{dixon2005synthesising}.

\subsubsection{Quantitative Synthesis: Meta-Analysis}

Meta-Analysis (or a integrative review) is a cornerstone in the quantitative synthesis landscape \cite{Gurevitch2018Metaanalysis}, and exemplifies the methodological rigor required to aggregate data across studies. Despite its potential for generating robust evidence, the meta-analytical process is fraught with challenges \cite{Haddaway2018Metaanalysis}. These include heterogeneity among study outcomes, variations in study quality, and the potential for publication bias, all of which necessitate sophisticated statistical techniques and critical judgment. The method's reliance on homogeneity and the need for comparable effect sizes across studies further complicate its application, especially in fields characterized by methodological diversity \cite{Metelli2020Challenges}.

\subsubsection{Qualitative Synthesis: Meta-Synthesis}

Meta-Synthesis, or an interpretive review, is the the qualitative counterpart that involves synthesizing findings from qualitative studies to uncover overarching themes or conceptual insights \cite{Xu2008Methodological}. This process is inherently interpretative, requiring a deep engagement with the textual data and an ability to discern patterns and meanings across disparate studies. The challenges here lie in maintaining the integrity and context of individual studies while seeking commonalities or divergent themes across the body of evidence \cite{Mohammed2016Metasynthesis}. The subjective nature of this synthesis approach underscores the need for transparency and reflexivity in the analytical process, ensuring that the synthesis is both comprehensive and faithful to the original studies \cite{Zimmer2006Qualitative}. The synthesis phase, irrespective of the methodological approach, is inherently complex due to the need to balance between the depth of individual studies and the breadth of the review's scope \cite{Hong2017Convergent}. This balancing act is further complicated by the varying quality of the studies, the diversity of their contexts, and the potential for conflicting findings \cite{Brunton2020Innovations}. As such, the synthesis not only requires methodological expertise but also a creative and critical approach to integrating knowledge \cite{Flemming2019Qualitative}. It is this phase that truly tests the reviewer's ability to construct a coherent narrative or theoretical framework that advances understanding in the field, making it one of the most intellectually demanding aspects of the SLR process \cite{Higgins2019Synthesising}.

\subsection{Current State of AI Tools and Research for Automation in SLRs}

Recent progress in AI, Natural Language Processing (NLP), and machine learning has significantly propelled the development of automation tools for supporting literature reviews. Techniques such as text mining and machine learning algorithms have demonstrated potential in automating various stages of the review process, from literature retrieval to data extraction \cite{Jonnalagadda2015}.  
These tools range from comprehensive research management platforms like Covidence\footnote{\url{http://www.covidence.org}}, ASReview\footnote{\url{https://www.asreview.nl}}, Mendeley\footnote{\url{https://www.mendeley.com}}, and Zotero\footnote{\url{https://www.zotero.org}}, to specialized tools aimed at specific literature review stages, such as 'connected papers'\footnote{\url{https://www.connectedpapers.com/}} for literature search and Elicit\footnote{\url{https://elicit.com/}} for individual paper analysis. Despite the growing availability of these tools, their adoption remains limited, with researchers citing challenges such as steep learning curves and inadequate support \cite{van2019usage}. More advanced AI tools provide citation support like Scite \footnote{https://scite.ai/}, as well as ScholarAI \footnote{https://scholarai.io/} integration tools with GPT models with numerous features for literature search support.

Meanwhile the body of research dedicated to automating SLR tasks is expanding, with several review papers categorizing these efforts \cite{feng2017text, muller2022application, van2021automation, de2023artificial, sundaram2023automating, da2021roadmap, wagner2022artificial, tsafnat2014systematic, marshall2019toward, atkinson2023cheap,bolanos2024artificial}. Nonetheless, the focus on automating the knowledge synthesis phase of SLRs remains minimal \cite{bolanos2024artificial} which is a reflection of the task's inherent complexity. Most research in SLR automation employs NLP and Information Retrieval techniques, with tasks such as data extraction and monitoring often framed as text summarization challenges \cite{bui2016extractive, altmami2022automatic}. 

\subsubsection{LLMs for research and SLRs automation}
\label{sec:llms_SLRs}

LLMs like GPT-3, have brought about transformative possibilities in SLRs. Rooted in the Transformer architecture, these models are well-suited for processing and synthesizing information from large corpora \cite{vaswani2017attention} that include academic texts. The fine-tuning of LLMs with domain-specific data presents an opportunity to enhance their effectiveness in generating accurate study summaries, albeit challenges in reliability, validity, and interpretability remain \cite{brown2020language}.

With the popularity of LLMs such as OpenAI's ChatGPT, Google's Gemini and Anthropic's Claude, SLR researchers now have a natural tendency to explore the capabilities of these models for SLR development. For example,~\citet{gupta2023utilization} used ChatGPT to generate novel SLR ideas and noted that the software was highly accurate.~\citet{hill2023methods} used Microsoft's Bing AI to extract study characteristics from research papers as data items.~\citet{castillo2023leveraging} compared six AI based tools ( (Forefront, GetGPT, ThebAI, Claude, Bard, and H2O) for the screening stage of SLR.~\citet{kumar2023analysis} used ChatGPT to generate small reports on research topics, but noted that the output lacks the rigour expected in academic writing.~\citet{zimmermann2024leveraging} used ChatGPt to answer several questions based on the title and abstract of research papers and reported that ChatGPT has an accuracy about 70\%.~\cite{alshami2023harnessing} used ChatGPT to automate several stages of SLR development process, including generating relevant keywords and phrases
for literature search, screening and filtering of
studies, and extracting and
synthesizing information from research papers. Interestingly, in the synthesizing step, they queried ChatGPT on individual papers, as well as multiple papers. However, they do not report quantifiable results on the performance of ChatGPT.~\citet{najafali2023truth} conducted even more bolder experiments by generating an entire SLR using ChatGPT, however noted that the output has many issues.

\subsubsection{Hallucination in LLMs and Factuality in SLRs}

The phenomenon of hallucination in LLMs, characterized by the generation of false yet plausible information, presents a critical challenge in the deployment of LLMs \cite{kalai2023calibrated}, with natural relevance for SLRs. This challenge is particularly acute given the stringent requirements for accuracy and reliability in SLR research, where any deviation from factual correctness can significantly undermine the integrity of the review process and its outcomes. Hallucinations can generally be categorized into two types, namely,  open-domain and closed domain hallucinations \cite{kalai2023calibrated}.

Hallucinations in open-domain contexts emerge when LLMs produce outputs not directly anchored in specific instances of the training data. This type of hallucination is particularly problematic in the context of SLRs, where the veracity of information is paramount. Strategies to mitigate open-domain hallucinations include enhancing the diversity and representativeness of training datasets, incorporating mechanisms for better context comprehension, and developing techniques for the LLM to recognize and flag potential inaccuracies. Meanwhile, closed-domain hallucinations occur within specific contexts, such as when LLMs generate content that diverges from the source text in tasks like translation or summarization. For SLRs, where synthesis of existing literature must adhere closely to the original texts, such hallucinations clearly pose significant risks. 

Reasons for LLMs generating hallucinations is commonly attributed to the presence of false information within the training data itself \cite{lin2022truthfulqa} or to the outdated or temporal nature of the training data which lacks information on recent developments \cite{aksitov2023characterizing}. However, these are not the sole contributors to hallucinations. Another factor is the LLMs' training approach to generate tokens sequentially, which can result in generating realistic but ultimately incorrect sequences of text \cite{zhang2023language}. Recently, \cite{kalai2023calibrated} demonstrated that LLMs inherently produce incorrect facts due to statistical principles rather than their design or data quality, with such errors often corresponding to the rate at which certain unique facts appear in the training data.

\subsection{RAG for Enhanced Factual Accuracy in SLRs}

Retrieval-Augmented Generation (RAG) \cite{Lewis2020RetrievalAugmented} is a framework that combines the capabilities of large language models (LLMs) with external knowledge sources through a retrieval mechanism. Unlike traditional language models that generate text based solely on the text's internal representation, RAG models retrieve relevant information from a knowledge base (like a database or the internet) and integrate this into the generation process. The core components of RAG include the retrieval mechanism, which fetches relevant documents or data, and the generative model, which synthesizes the retrieved information into coherent and contextually relevant responses. The theoretical principles underpinning RAG stem from the need to enhance language models with the ability to access and utilize external, structured knowledge \cite{asai2023SelfRAG:}. This is in response to the limitations of traditional LLMs that rely solely on their pre-trained parameters for knowledge, which can be outdated or incomplete. The typical architecture of RAG systems involves a retriever that fetches relevant information from a database and a generator that incorporates this information into the final output, thus, the integration of these components allows the model to produce contextually enriched and factually accurate text \cite{Lewis2020RetrievalAugmented}. 
The incorporation of RAG into the workflow of LLMs, therefore, presents a sophisticated approach to augmenting the model's knowledge base beyond its pretraining, specifically tailored to the demands of the evolving nature of studies in SLRs. By allowing the model to access an external corpus of domain-specific literature the context within which the LLM operates, it attains the ability to be enriched while a critical countermeasure to the model's propensity for generating plausible yet factually incorrect information - hallucination - is mitigated. In the realm of SLRs, where the precision of synthesized knowledge is paramount, RAG's ability to draws upon relevant information from a targeted corpus helps ensure that the generative outputs of LLMs are anchored in verifiable data. 

\subsection{Advancing LLMs for Specialized Domains: From Pretraining to Fine-Tuning}

In training LLMs, they initially undergo pretraining on extensive, diverse datasets, acquiring a foundational grasp of both language and knowledge. This stage equips LLMs with a generalist understanding, however, the leap from generic text generation to specialized domain proficiency requires additional training in the form of fine-tuning which is a process where LLMs are further trained on relevant datasets to align their outputs with new requirements \cite{Gururangan2020}.

\subsubsection{PEFT: A Paradigm Shift in Finetuning}

The evolution of fine-tuning practices, particularly with the advent of Parameter-Efficient Fine-Tuning (PEFT) techniques \cite{howard2018universal}, marks a significant shift towards more sustainable and effective model optimization. PEFT, focuses on updating a selective subset of the model's parameters. This approach contrasts with conventional fine-tuning, where a substantial portion of the parameters is modified, leading to high computational costs. By concentrating updates on strategically chosen parameters, PEFT techniques enhance the model's performance on domain-specific tasks without the need for extensive computational resources, thus making fine-tuning feasible even for the most expansive LLMs \cite{houlsby2019parameter, dettmers2023qlora}. The integration of PEFT techniques in the fine-tuning process is instrumental in tailoring LLMs to the narrow and specific knowledge requirements fo chosen domains such as those encountered in  (SLRs), which would not be captured by the pretrained process of LLMs.

Among the myriad of PEFT techniques, LoRA \cite{hu2021lora} and more recently NEFTune \cite{jain2023neftune}, stand out for their contributions to enhancing LLMs' capabilities. LoRA (Low-Rank Adaptation), refines the model's efficiency by optimizing a small, critical subset of parameters, thus minimizing the computational overhead while maintaining, or even enhancing, the model's performance in domain-specific applications \cite{dettmers2023qlora}. NEFTune on the other hand introduces an innovative approach by incorporating random noise into the embedding vectors during fine-tuning, thereby improving the model's ability to follow instructions and engage in meaningful conversations by enhancing its generalization capacity \cite{jain2023neftune}.

\subsection{Literature summary and gaps}

The literature review underscores the potential and gaps in the intersection of SLRs and LLMs. While advancements in text mining and machine learning herald efficiency in literature retrieval and screening, the synthesis phase remains a challenge that has not been sufficiently addressed, and made difficult by the phenomenon of LLM hallucinations and the inability to audit the source information from the LLMs' responses. The review reveals the potential of fine-tuning LLMs with SLR-specific datasets to tailor these models for more accurate knowledge synthesis. However, work is required to ensure that both fidelity of synthesized knowledge and auditing of LLM responses can be achieved. Against this backdrop, this paper proposes an SLR-automation Framework to bridge this gap by leveraging domain-specific fine-tuning of LLMs via the most recent advances in PEFT, on selected SLR papers and targetting the knowledge synthesis stage of SLRs, together with novelties for embedding fidelity and mitigating hallucination in the LLM responses. The viability of the proposed framework is explored and illustrated using a case study that will seek to replicate a PRISMA-conforming SLR as a ground truth against which the proposed framework can be evaluated.

Based on the research gaps in the literature, this study attempts to answer the following research questions:

\begin{itemize}
    \item (RQ1) Can we leverage LLMs in combination with finetuning and/or RAG in order to facilitate the synthesis phase of an SLR process?
    \item (RQ2) How can the extraction of finetuning datasets from an SLR corpus be effectively automated?
    \item (RQ3) Can finetuning of LLMs be effectively conducted on relatively small datasets representing a very narrow domain?
    \item (RQ4) Can we achieve high fidelity of LLM responses and ensure that we can audit and verify the source information of LLM responses, and what evaluation metrics are suitable?
    \item (RQ5) Are the proposed SLR-automation approaches viable for replicating a previously published SLR study?

\end{itemize}


\section{Proposed SLR-automation Framework}

\begin{figure}[hbt]
    \centering
    \includegraphics[width=.8\textwidth]{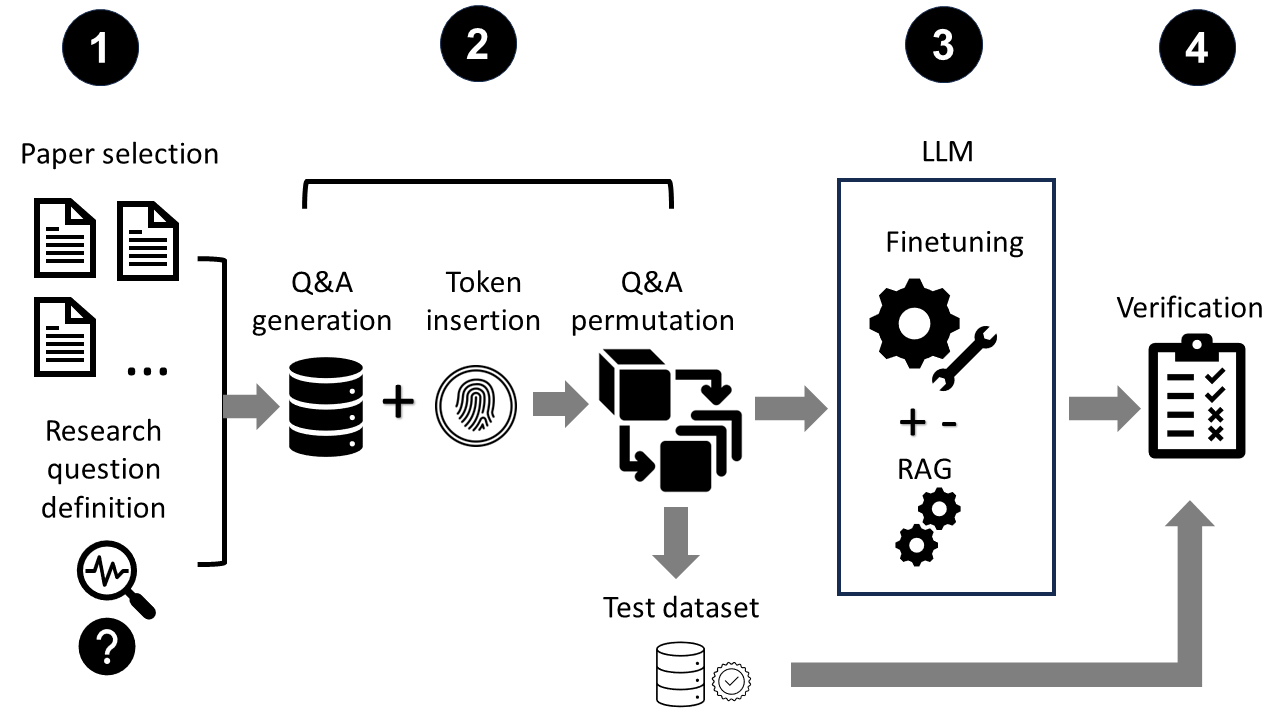}
    \caption{Overview of the proposed SLR-automation Framework for evidence and knowledge synthesis.}
    \label{fig:diagram1}
\end{figure}

Our proposed framework for automating SLRs is a sequential process designed to extract a dataset that encapsulates necessary information from SLR-selected papers for the purpose of finetuning LLMs in order to facilitate the knowledge synthesis phase of the SLR process. Figure \ref{fig:diagram1} outlines the proposed four-step process.

\subsection{Step 1 - Paper Selection and Research Question Definition}
The framework initiates with a dual focus on SLR paper selection and research question articulation which is covered by the planning, selection and extraction stages of the SLR process (Figure \ref{fig:diagramSLR} Steps 1 to 3). Once the target papers for an SLR are identified and the overarching and supporting research questions for the study have been defined, they both then serve as inputs to the next Step 2, and direct the automation process of data extraction and processing. 

\subsection{Step 2a - Automated Q\&A Data Extraction and Synthesis}

In this step, we developed a program \footnote{The tool is publicly available here: https://github.com/peterjhwang/slr-helper} that leverages the downstream summarization as well as the question and answering capabilities of an advanced LLM to extract a dataset from all the papers. This program systematically processes each selected paper, using the chosen LLM's summarization capabilities to extract the core content of primary paper sections and subsections into concise summaries alongside the papers' meta-data. From these distilled summaries, the program then uses an LLM to  automatically generate a series of domain-relevant questions that are common denominators and can be answered by all studies. This set of questions is then 
used to interrogate each paper and extract answers via a chosen LLM, and the combination of the two forms question and answer (Q\&A) pairs that represent our finetuning dataset. Simultaneously, we enable the customization of domain-specific, researcher-defined questions informed and motivated by the objectives and aims of an SLR being undertaken. These questions are tailored to the unique contours of the study's domain and are formulated to draw out information that is particularly relevant for the SLR's purposes. This dual strategy ensures that while our approach remains systematically consistent across the majority of academic disciplines, it is also flexible enough to accommodate the specialized requirements of the academic field under investigation.

With both sets of questions, we are then able to automatically generate \textit{paper-level} Q\&A pairs that encapsulate the insights of individual studies at different levels of detail and granularity. To achieve a synthesis that spans the corpus of all studies under investigation, we further collate key summaries from across all papers and from these, generate additional \textit{SLR-level} Q\&A pairs. These pairs are designed to provide responses that are relevant across the entire body of selected works, effectively offering a synthesized narrative that captures the collective information of the SLR papers. 
Through this process, we ensure that the resulting Q\&A dataset is not only rich in detail and precision at the level of individual papers but also possesses the breadth and depth required to reflect the evidence synthesis of a systematic literature review.

\subsection{Step 2b - Token Insertion as New Knowledge Markers}

The incorporation of distinct \textit{knowledge markers}, referred to as unique tokens, into the fine-tuning process of LLMs is a theoretical construct designed to create explicit linkages between the model's augmented neural network parameters and the new knowledge encapsulated from the SLR papers. The motivation behind this approach is to achieve a higher degree of fidelity  and precision in the model's output during inference. When an LLM is fine-tuned with additional data, it assimilates this new information by adjusting a small percentage of its existing parameters while adding a larger set of new parameters. Without a mechanism to distinguish between pre-existing knowledge acquired from pretraining and newly acquired information, the model risks providing responses based on its pretraining rather than the domain-specific insights gained from the subsequent fine-tuning step. Since it is imperative that during inference time, the model does not access facts from its knowledge base developed during pretraining, and instead, only responds with facts acquired during finetuning, an explicit mechanism is needed to ensure this occurs. Without such mechanisms, the results of the SLR risk being contaminated with knowledge that is outside of the scope of the selected papers, thus, invalidating the research. In essence, the aim of the proposed approach is to insulate the LLMs responses from its pretrained knowledge but to nonetheless leverage its natural language capabilities that it gained during pretraining. 

Therefore, these tokens or markers serve as signposts within the neural network, directing the model to access and utilize the parameters associated with the SLR data and facts when generating responses. It also enables the model to explicitly state the source information for its responses so that the provenance of information can be audited. 
This approach provides a method for researchers to verify that the output is indeed based on the recent fine-tuning, enabling a level of interpretability and trust in the model's responses that is critical for academic rigor and reliability and represents a strategic advancement in using LLMs for SLR.

\subsection{Step 2c - Q\&A Permutation}

As part of Step 2, to ensure that the LLMs can handle a variety of linguistic expressions during inference time, we introduce permutations in the Q\&A pairs. For each original question, we create several semantically equivalent variations. Since accurate recall of facts is the primary objective of the system rather than style, linguistic quality, and variability of the text, we retain the same answer for each permutation of the corresponding questions. This step is critical for training the LLMs to recognize and correctly respond to different phrasings of the same question which different users are likely to generate. 

\subsection{Step 3 - Finetuning and RAG}

In the third step of our SLR-automation framework, we propose specific finetuning methods for LLMs, complemented by the integration of RAG processes. To enhance the finetuning efficiency, we propose employing PEFT techniques \cite{howard2018universal}, which modify only a subset of the model's parameters. This is crucial, as fine-tuning all parameters of large LLMs is computationally prohibitive and may not yield proportionate improvements in performance. PEFT techniques offer a balance, concentrating the model's learning capacity on those parameters that most significantly influence the domain-specific tasks, thereby optimizing both resource expenditure and model proficiency. RAG, on the other hand, serves as an augmentative mechanism during inference. It dynamically retrieves and incorporates information from the SLR corpus into the LLM's generative process, potentially providing an additional layer of contextual support. This optional step may ensure that the model's output is not solely reliant on internalized patterns but is also substantiated by direct data references, thereby potentially enriching the depth and factual accuracy of the synthesized content and responses.

\subsection{Step 4 - Verification}
The final phase of the proposed framework tests the framework to ensure that it has the ability to accurately generate responses with respect to the selected SLR papers. The test dataset is extracted from the main Q\&A finetuning dataset and withheld from the finetuning process. An automated process is subsequently initiated after finetuning is completed which tests the LLMs ability to answer the selected questions correctly using metrics that are specifically designed for factual validation.

\section{Methodology}

The methodology covers details of the chosen Gold Standard SLR as a use case for replication and validation of the proposed framework. It compasses details on the dataset extraction and preparation for LLM fine-tuning, the experimental design outlining the setup of LLM technologies, selection of LLM types, detailed RAG implementation, finetuning hyperparameters, and hardware setup. The evaluation measures are discussed which cover quantitative metrics and qualitative analyses to assess the information recall.

\subsection{Use Case: Replication of an Existing SLR}

We selected a peer-reviewed and published PRISMA-SLR within the domain of learning analytics as the Gold Standard, with the goal of investigating whether the proposed SLR-automation framework  has the capability to generate responses from which a  manually conducted SLR can be replicated.
The target SLR paper is titled \enquote{Learning Analytics Dashboard: A Tool for Providing Actionable Insights to Learners} \cite{susnjak2022learning}. This SLR is contemporary, being aligned with the recent advancements in learning analytics dashboards, and was chosen for its adherence to the PRISMA guidelines as well as due to authors' familiarity with the work. The SLR itself focused on aspects of advancements in educational technology, delving into the impacts of learning analytics dashboards (LADs) on university student outcomes, scrutinizing the technological underpinnings, the analytics methodologies most commonly employed, and the demographics of the study cohorts used in trials. Furthermore, it examined the existing gaps and posited future trajectories for research in this field. This SLR encompassed 17 papers. Its selection was strategic not only because it aligned with the expertise of this study's authors which assisted in guaranteeing the veracity of LLM responses, but also offered an insight into the viability and effectiveness of finetuning LLMs on relatively small datasets. 

\subsection{Automated Dataset Extraction and Preparation}

Step 2 of the proposed framework in Figure \ref{fig:diagram1} was fully automated in order to generate a finetuning dataset. A software tool was built using Python to systematically parse the 17 PDF files representing the SLR papers and generate Q\&A pairs for the finetuning dataset. To begin with, this tool extracted the metadata from each paper via OpenAI's GPT-4 API calls and before data generation commenced, it created a high-level summary of each paper. The summaries of each paper were then collated, and GPT-4 was tasked with generating a set of questions that were relevant across all the papers. This produced a vehicle for automatically extracting domain-specific information that was common across all the studies. These questions were then augmented with a further set of \textit{researcher-defined} questions that reflected the narrower aims of the SLR study and were closely aligned to the objectives and overarching questions of the SLR being conducted. Once this broad set of questions was collected, GPT-4 was used via API calls to generate answers to each one from every study. These initial Q\&A pairs we refer to as \textit{paper-summary-level} data points. These data points captured high-level paper summaries that were relevant to both the research domain and the aims of the SLR study.

The next phase involved the generation of another set of Q\&A data points based on data from individual studies at different levels of granularity. In this phase, our processing divided each paper into chunks\footnote{The chunks usually comprised key sections of each paper.}, and asked GPT-4 to directly generate Q\&A pairs based on the content of each chunk. This method yielded another dimension of information from each paper that reflected some more distinct aspects of each study alongside, like their expanded and more detailed contributions, idiosyncratic methodological details, key findings, and future research directions. This process was then repeated with even smaller chunks of text which roughly represented individual paper paragraphs. This more detailed approach allowed for the extraction of even more granular information from each study in the form of Q\&A pairs, revealing yet finer details of the methodologies, analyses, and discussions presented in the studies. By dissecting the papers at a paragraph level, this method augmented the depth of the overall finetuning dataset with a more detailed set of Q\&A pairs. This layer, we collectively refer to as \textit{paper-level} Q\&A pairs that offer a more granular insight into each study compared to \textit{paper-summary-level} data points.

While the above strategy generated data points that would enable researchers to \enquote{talk to} the content of each paper, the subsequent strategy aimed to allow researchers to \enquote{talk across} all the studies simultaneously. These Q\&A pairs we refer to as \textit{SLR-level} data points which served the with the aim of facilitating the answering of questions that relate to the synthesis of knowledge across all the studies under investigation. From the implementation perspective, this second layer of data extraction involved collating answers from each paper that were generated from questions used for creating \textit{paper-summary-level} Q\&A pairs. Then, Q\&A pairs were generated via GPT-4 based on all the collated answers.
By combining paper and SLR-level Q\&A pairs, our finetuning dataset not only facilitated equipping LLMs with in-depth facts from singular studies, but also attempted to make possible LLM prompts that support queries requiring a comprehensive synthesis of knowledge and evidence across the entire SLR corpus, achieving both the breadth and depth needed to support a systematic study of a corpus of literature and to answer specific research questions.
 
The initial training dataset following the data extraction and Q\&A pairs generation comprised 1503 samples and of these, approximately 28\% were paper-summary-level, 49\% were paper-level extracted from larger chunks, 31\% were paper-level extracted from paragraphs, and 2\% from the SLR-level. To augment the robustness of the finetuned LLMs during inference, we implemented a novel refinement step: for each original question, we generated ten alternative formulations. These variants were designed to be semantically diverse yet structurally aligned, ensuring that despite the variability in phrasing, the essence of the questions remained intact, eliciting the same precise response which resulted in producing a final training dataset of 16423 samples. This strategic augmentation served a dual purpose. Firstly, it aimed to significantly enhance the model's ability to comprehend and respond accurately to a broader spectrum of question formulations from different users, mirroring the diverse linguistic styles encountered in real-world academic discourse. Secondly, it aimed at increasing the size of the dataset. For the test dataset, we generated three additional permutations for each unique question to arrive at a final dataset of 4962 data points, where the distribution of paper-level and SLR-level questions was the same as that in the training dataset. Table \ref{tab:questions_answers} demonstrates an example of a text summary (answer), for which 11 different but semantically equivalent questions were generated using GPT-4 for the training dataset and an additional three for the test dataset.

{\fontsize{8pt}{10pt}\selectfont
\begin{table}[ht]
\centering
\begin{tabular}{>{\raggedright\arraybackslash}p{0.2\linewidth}>{\raggedright\arraybackslash}p{0.7\linewidth}}
\toprule
\textbf{Answer} & \textbf{Questions} \\ 
\midrule
\fontsize{8pt}{10pt}\selectfont
\multirow{15}{=}{The paper's findings inform future directions for research by highlighting the importance of personalizing LADs to course requirements and the potential of student-centered dashboards to enhance engagement and performance. It suggests further exploration of LADs' role in supporting self-regulated learning and the challenges of personalization.} 
  & \fontsize{8pt}{10pt}\selectfont\textbf{Training Questions} \\
  \cline{2-2}
 & \fontsize{8pt}{10pt}\selectfont1. How do the results of the study shape future research and development paths in learning analytics and educational technology? \\ 
 & \fontsize{8pt}{10pt}\selectfont2. What implications do the paper's outcomes have for the future of research and development in learning analytics and educational technology? \\ 
 & \fontsize{8pt}{10pt}\selectfont3. How can the findings from the paper influence the forthcoming trends in research and development for learning analytics and educational technology? \\ 
 & \fontsize{8pt}{10pt}\selectfont4. What role do the paper's findings play in setting the course for future research and development in learning analytics and educational technology? \\ 
 & \fontsize{8pt}{10pt}\selectfont5. How will the paper's discoveries impact the trajectory of research and development in learning analytics and educational technology moving forward? \\ 
 & \fontsize{8pt}{10pt}\selectfont6. What effect do the paper's conclusions have on the future strategies for research and development in the domains of learning analytics and educational technology? \\ 
 & \fontsize{8pt}{10pt}\selectfont7. How are the paper's findings expected to steer future research and development initiatives in learning analytics and educational technology? \\ 
 & \fontsize{8pt}{10pt}\selectfont8. In what ways might the paper's results influence upcoming research and development activities in the field of learning analytics and educational technology? \\ 
 & \fontsize{8pt}{10pt}\selectfont9. In what ways do the conclusions of the paper guide the next steps for research and development within the realms of learning analytics and educational technology? \\ 
  & \fontsize{8pt}{10pt}\selectfont10. In what manner do the insights from the paper direct future research and development efforts in the field of learning analytics and educational technology? \\ 
  & \fontsize{8pt}{10pt}\selectfont11. How do the paper's findings inform future directions for research and development in the field of learning analytics and educational technology?
 \\  
  & \fontsize{8pt}{10pt}\selectfont\textbf{Test Questions} \\
  \cline{2-2} 

& \fontsize{8pt}{10pt}\selectfont12. In what capacity do the study's findings shape the forthcoming strategies for research and development across learning analytics and educational technology? \\
& \fontsize{8pt}{10pt}\selectfont13. How might the conclusions drawn from the paper steer the future of innovation and inquiry in the spheres of learning analytics and educational technology? \\
& \fontsize{8pt}{10pt}\selectfont14. What influence do the outcomes of the paper exert on the trajectory of upcoming investigations and advancements in learning analytics and educational technology? \\

 \bottomrule
\end{tabular}
\caption{Example of the dataset permutation step for questions relating to an answer from one specific paper in the target SLR study.}
\label{tab:questions_answers}
\end{table}
}

\subsubsection{Dataset Format and Characteristics}

As part of the dataset processing in Step 2 in the proposed framwork in Figure \ref{fig:diagram1}, a specific format to the structure of the Q\&A pairs was enforced, and unique tokens were inserted into each Q\&A data point. 
For clarity, an illustrative example of a Q\&A data point from the dataset can be seen as follows:

{\fontsize{8pt}{10pt}\selectfont
\begin{verbatim}{
    "instruction": "According to the 2023SLR dataset, in the aljohani2019integrated paper, 
                    how does the paper conceptualize Learning Analytics Dashboards (LADs)
                    within the framework of digital learning environments?",
    "output": "In the data used for the 2023SLR, the paper conceptualizes Learning
               Analytics Dashboards (LADs) as a knowledge and learning management oriented 
               infrastructure for the end-to-end management and consumption of educational 
               data within digital learning environments. It emphasizes the integration of 
               student-centered data as a critical challenge for personalized curricula and 
               learning experiences. Source: aljohani2019integrated"
}
\end{verbatim}
}

This data point exemplifies the dataset's innovative use of unique tokens included in the post-processing step that embeds specific knowledge markers within the fine-tuning dataset which is crucial for enabling the model's capacity to differentiate between newly acquired information and its pre-existing knowledge base from pretraining. The example token `2023SLR` serves as a \textit{corpus-wide} identifier, linking this and all the data points to the broader SLR dataset, while in this example, the `aljohani2019integrated` token operates as a granular paper-level, tethering the Q\&A pair to a particular study within the corpus. The inclusion of a source citation at the end of each output (`Source: aljohani2019integrated`) is a deliberate design choice, ensuring that each response generated by the LLM during inference can be sourced and traced back to its originating study and verified if needed, thus enhances the accountability of the model's outputs.

\subsection{LLM Selection and Finetuning Strategy}

In the experimentation of the proposed framework, the Mistral-7B base model \cite{jiang2023mistral} was selected as the foundation for fine-tuning due to its efficiency and high benchmark performance, particularly in English language tasks. This 7.3 billion parameter model leverages advanced techniques such as Grouped-query attention and Sliding Window Attention, enabling it to outperform larger models across a variety of benchmarks, including commonsense reasoning, making it an ideal candidate for fine-tuning to suit our specific SLR automation needs.
Conversely, the Mistral-7B-Instruct variant, already fine-tuned for instructional tasks, was employed in conjunction with the RAG approach to leverage its capabilities in handling Q\&A instructional content. The Instruct model's demonstrable adaptability and performance on tasks like MT-Bench, coupled with its ability to generalize across tasks, provided an appropriate candidate for generating accurate and contextually relevant responses in our SLR framework experimentation. 

\subsubsection{Finetuning implementation}

In the fine-tuning of the LoRa and NEFTune LLM models we focused on calibrating a selection of the hyperparameters that are presented in Table \ref{tab:hyperparametersComparison}, with all others set to their default values.

\begin{table}[h]
\centering
\small
\begin{tabular}{rll}
\hline
\textbf{Parameter} & \textbf{NEFTune} & \textbf{LoRA} \\ \hline
\texttt{learning\_rate} & 5e-05 & 5e-05 \\
\texttt{num\_train\_epochs} & 150 & 150 \\
\texttt{per\_device\_train\_batch\_size} & 4 & 4 \\
\texttt{per\_device\_eval\_batch\_size} & 4 & 4 \\
\texttt{fp16} & Enabled (true) & Enabled (true) \\ \hline
\end{tabular}
\caption{Comparison of Hyperparameters for NEFTune and LoRA Model Fine-Tuning}
\label{tab:hyperparametersComparison}
\end{table}

\subsection{RAG implementation details}

In our study, we implemented two RAG systems, leveraging the Weaviate vector database (version 1.23.9) and the OpenAI Embedding model `text-embedding-3-large`\footnote{This embedding model was notable for its efficiency and ranked 7th on the HuggingFace leaderboard at the time of writing this paper.} for pre-retrieval processes.  For the post-retrieval stage, the initial model used was OpenAI's `gpt-3.5-turbo`.
The pre-retrieval phase involved optimizing the indexing process by chunking data using LLMs, focusing on contextually relevant segments. The vector database was structured with the attributes `text` and `source`, where `text` was searchable and `source` was filterable. 
During the retrieval phase, we utilized Langchain's self-querying retriever for its simplicity\footnote{Future implementation intentions are to upgrade to a hybrid search model for enhanced accuracy and more refined filtering mechanisms.}.
In the post-retrieval process, we employed an aggregation prompt template designed to integrate the retrieved context into a coherent and concise answer, limited to three sentences for any given question.

\subsection{Experimental Design}

We experimented with five different combinations of finetuning and RAG in order to determine what the most effective approaches could be for the SLR-automation process. Our experimental setup investigated the following methodologies:

\begin{enumerate}
    \item Baseline: Evaluate the Mistral-7B-Instruct on its knowledge and ability to answer SLR-related test dataset questions.    
    \item Fine-tuning LLMs using LoRA: Leveraging Low-Rank Adaptation for fast and efficient parameter adjustment.
    \item Fine-tuning LLMs using NEFTune: Introducing noise into embedding vectors to investigate effects on generalization improvements.
    \item Instruct LLM  + RAG with Raw Articles: Combining LLMs with Retrieval-Augmented Generation, using unprocessed article text as the retrieval corpus.
    \item Instruct LLM + RAG with Auto-Extracted Data: Employing RAG with a knowledge base of automatically extracted data comprising the finetuning dataset for focused information retrieval.
    \item Best Finetuned LLMs + Best RAG Solution: Integrating the top-performing fine-tuning and RAG methods to optimize SLR automation.
\end{enumerate}

Each method also summarised in Table \ref{table:experimental_methods}, was evaluated for factually correct answers with respect to the SLR dataset.

\begin{table}[ht]
\centering
\fontsize{8pt}{10pt}\selectfont
\begin{tabular}{l l}
\toprule
\textbf{Method} & \textbf{Description} \\
\midrule
Baseline Mistral-7B-Instruct & Establish the degree of pretraining and instruct-tuning knowledge within the model \\
LoRA Finetuning & Fine-tuning with Low-Rank Adaptation \\
NEFTune Finetuning & Fine-tuning with Noise-Enhanced Fine-Tuning \\
Instruct LLM + RAG (Raw) & Retrieval-Augmented Generation with raw articles \\
Instruct LLM + RAG (Extracted) & RAG with auto-extracted data \\
Best Combination & Best fine-tuning + RAG approach \\
\bottomrule
\end{tabular}
\caption{Summary of Experimental Methods for SLR Automation}
\label{table:experimental_methods}
\end{table}

\subsection{Hardware specifications and training runtimes}

Fine-tuning and RAG execution were performed on a Linux system powered by an AMD EPYC 7713 64-bit 8-Core Processor with 259GB of RAM. For the training, we utilized an NVIDIA A40 GPU, which is built on an 8 nm process and equipped with 48 GB of GDDR6 memory. This GPU features 10,752 shading units, 336 tensor cores, and 84 ray tracing cores, supporting a base clock speed of 1305 MHz and a boost clock up to 1740 MHz. The total power consumption of the GPU is 300 W. The training processes for both NEFTune and LoRA models took approximately 70 hours to complete 150 epochs each.

\subsection{Evaluation}

Our evaluation necessarily centered around assessing the factual accuracy of responses generated by the various methodologies applied to the SLR dataset. This assessment was conducted through a combination of quantitative and qualitative analyses, utilizing test sets.

\subsubsection{Test Dataset}

For the quantitative analysis, we employed an 80/20 train/test split. The qualitative test set comprised 11 gold standard answers, which were derived from key SLR-level findings across the 17 papers within the target SLR paper. These findings represented the core contributions of the SLR study and were articulated into Q\&A pairs. 

\subsubsection{Quantitative Metrics}

The quantitative evaluation was grounded in two metrics, each devised for its relevance to the evaluation of factual accuracy in the context of SLR automation. The first was FEVER \cite{thorne2018FEVER}, whose usage is consistent with other studies \cite{yao2022ReAct,bekoulis2020A} concerning factual text evaluations. The next was a variation on FEVER which we refer to as the Consistency Grading Scale (CGS) that offers a more granular analysis of the alignment between response texts and reference materials.

\begin{enumerate}
\item \textbf{FEVER (Fact Extraction and VERification)}: The FEVER metric was a key evaluation framework used in conjunction with GPT-4 to automate the testing. FEVER scoring is designed to assess the factual accuracy of generated responses by cross-referencing the LLM-generated responses against ground truth responses which were both submitted to GPT-4 to assess. The complete GPT-4 prompt can be seen in Appendix  \ref{fever_eval_promt}.  FEVER scoring determines the veracity of the claims, categorizing them into distinct labels for clarity. The labels used in the standard FEVER evaluation are:
\begin{itemize}
    \item \textit{UPPORTED }: The claim is directly corroborated by the evidence in the ground truth test set.
    \item \textit{REFUTED}: The claim is directly contradicted by the evidence in the ground truth test set.
    \item \textit{NOT ENOUGH INFO}: There is insufficient evidence in the ground truth test set to either support or refute the claim.
\end{itemize}

\setcounter{enumi}{1}
\item \textbf{Consistency Grading Scale (CGS)}: Building upon the foundational principles of FEVER, CGS introduces a more granular continuum for evaluating the fidelity of information presented by LLM responses in the context of systematic literature reviews (SLRs). The CGS was also used in conjunction with GPT-4 in order to derive the assessments, where the complete prompt can be seen in Appendix \ref{q_eval_promt}.
This grading scale quantitatively assesses the alignment of generated responses with verified reference texts, offering a spectrum of consistency levels ranging from complete contradiction to full support. The CGS is defined as follows:
\begin{itemize}
\item \textit{Fully Consistent (2)}: The claim in the response text is unambiguously corroborated by the evidence in the ground truth, with complete agreement in the source information.
\item \textit{Partially Consistent (1)}: The response text is generally aligned with the reference material; however, minor discrepancies or inadequacies in detail, accuracy, or source alignment are evident.
\item \textit{Not Enough Info (0)}: The available evidence is insufficient to substantiate or refute the claims made in the response text, indicating a gap in information.
\item \textit{Partially Contradictory (-1)}: While some aspects of the response text may align with the reference material, significant contradictions or factual inaccuracies are present.
\item \textit{Contradictory (-2)}: The response text is in direct and total opposition to the evidence presented in the ground truth, indicating clear factual inaccuracies.
\end{itemize}

\end{enumerate}

The FEVER evaluation comprises the calculation of the percentages of responses labelled as either \textsc{SUPPORTED}, \textsc{REFUTED}, or \textsc{NOT ENOUGH INFO}. Meanwhile, for the CGS evaluation, a mean is calculated across all the values across every response to derive a value that ranges from -2 (totally inaccurate) to 2 (perfectly accurate). GPT-4 was leveraged for automating the evaluations against the criteria defined above which is a consistent approach with previously published studies \cite{mcintosh2024inadequacy,mcintosh2023harnessing}. Prompts for both CGS and FEVER were empirically developed and refined by iterative testing against GPT-4 in order to ensure that correct grading responses were being generated. The prompts can be seen in Appendix \ref{evaluationprompts}. In order to establish the validity of the devised evaluation metrics, 100 samples from the test dataset were sampled and inter-rater reliability was analyzed between the responses of two human evaluators and GPT-4 for both CGS and FEVER metrics, whose results are presented in the subsequent section.

\subsubsection{Qualitative Analysis}

The qualitative component of our evaluation framework was designed to gauge the depth and relevance of the synthesis provided by the LLM-generated responses. An author from this study who is a domain expert and well-versed in the details of the subject matter was enlisted to perform a comprehensive review of each response. Their analysis was centered on determining if the response fell into one of the two categories: \textit{Supports} and \textit{Does Not Support}. These categories determined the extent to which the responses could encapsulate key synthesis information from the aggregated SLR studies which enable the core conclusions and findings of the benchmark SLR paper to be replicated.

\begin{itemize}
\item \textbf{Supports}: This category was reserved for responses that demonstrated an accurate presentation of  the SLR content, effectively summarizing all the key points, methodologies, results, and conclusions. The responses classified under this category were those that provided comprehensive synthesis information, enabling the researcher to completely replicate the key findings of the target SLR study as reported in the published paper.

\item \textbf{Does Not Support}: Responses falling into this category were those that failed to capture all the details necessary for the replication of the key findings of the target SLR study. These responses may include inaccuracies or omissions in key details.

\end{itemize}

Each response was manually evaluated against these categories, with a subject expert providing detailed feedback on the fidelity of the responses. To ensure the objectivity and reliability of the qualitative analysis, the gold standard Q\&A pairs used as benchmarks were kept confidential from the team responsible for the LLMs' development and training.

\section{Results}

Our analysis of results first considers the validity of the CGS and FEVER metrics, and then compares the performances of all the proposed approaches. A correlation analysis is then provided based on the responses of all methods, followed by a more detailed analysis of the best-performing approach.

\subsection{Validation of evaluation method}

We begin first by establishing the validity of the devised evaluation metrics. Tables \ref{table:cgs_correlation} and \ref{table:fever_correlation} show the correlation matrices analyzing the inter-rater reliability among the two independent human evaluators' (H1 and H2) scores as well as that of GPT-4, showing the utility and limitations of both CGS and FEVER within the context of factuality assessment.

\begin{table}[ht]
    \centering
    \begin{minipage}{0.45\textwidth}
        \centering
        \small
        \begin{tabular}{rccc}
            \hline
            Rater & GPT-4 & H1 & H2 \\ \hline\hline
            GPT-4 & 1.0 & & \\
            H1 & 0.647 & 1.0 & \\
            H2 & 0.739 & 0.669 & 1.0 \\ \hline
        \end{tabular}
        \caption{Inter-rater correlation matrix for the CGS metric.}
        \label{table:cgs_correlation}
    \end{minipage}
    \hfill
    \begin{minipage}{0.45\textwidth}
        \centering
        \begin{tabular}{rccc}
            \hline
            Rater & GPT-4 & H1 & H2 \\ \hline\hline
            GPT-4 & 1.0 & & \\
            H1 & 0.49 & 1.0 & \\
            H2 & 0.6 & 0.557 & 1.0 \\ \hline
        \end{tabular}
        \caption{Inter-rater correlation matrix for the FEVER metric.}
        \label{table:fever_correlation}
    \end{minipage}
\end{table}

The CGS metric displays moderate to strong correlations, particularly with the H2 rater exhibiting a robust 0.74, suggesting a substantial concurrence with the GPT-4 evaluations. H1 demonstrated a moderate agreement with a correlation of 0.65 with GPT-4. These figures are noteworthy, given the subjective nature inherent in qualitative evaluations. The inter-rater reliability, represented by a correlation of 0.67, although not indicating perfect agreement, nevertheless reveals a meaningful consensus between the human raters that overall affirms the CGS as a reliable tool within its application scope.

Contrastingly, the FEVER metric was characterized by lower correlation values and indicated a more modest agreement between evaluators – which is a reflection of the challenges faced in calibrating subjective analytical tools in general. With H1 at 0.49 and H2 at 0.6 correlation with GPT-4, there was discernible inter-evaluator consonance, albeit less pronounced than that within the CGS metric.
The inter-rater correlation of 0.56, while significant, underscored the need for further evaluator calibration to mitigate interpretive disparities. Upon a deeper investigation into the divergences, the results showed that there were significant differences in the manner in which the human raters evaluated the 'NOT ENOUGH INFO' category with respect to each other and GPT-4, with both human raters underrating this category and preferring either the \textsc{SUPPORTED} or \textsc{REFUTED} categories instead. Thus, while the FEVER metric offers meaningful insights into the veracity of claims for the analysis in this study, its application necessitates a more critical interpretation. For the purposes of this study, CGS can be relied upon to provide a broader spectrum of consistency in evaluation, and FEVER serving as a barometer for broader factual accuracy.

\subsection{Quantitative analysis accuracies}

The quantitative outcomes, encapsulated in Table~\ref{tab:fever}, show the performance across \textsc{REFUTED}, \textsc{NOT ENOUGH INFO}, and \textsc{SUPPORTED} categories, thereby offering a granular perspective on the factual alignment of each method. All results are rank-ordered with respect to the \textsc{SUPPORTED} category.  We first draw attention to the results of the Baseline method which represents responses from the unmodified Mistral-Instruct LLM and indicates the potential existence of any SLR-relevant prior knowledge arising from its pretraining and instruct-tuning processes, while also serving as a benchmark to determining the degree to which proposed approaches have succeeded in achieving an improvement. The baseline indicates the ability to correctly respond to 14.5\% of the test data, with the majority of the responses being categorised as \textsc{NOT ENOUGH INFO}.
We see next that the NEFTune method emerged as the most effective, with 89.2\% of its responses being \textsc{SUPPORTED}, indicative of its strong factual accuracy and alignment with the actual SLR corpus. LoRA, with an 87.7\% `SUPPORTED` rate, also demonstrated significant reliability, albeit marginally lower than NEFTune. In contrast, the integration of RAG with a fine-tuned dataset (RAG + FD) presents a discernible shift in distribution across all three categories, with a notable increase in `NOT ENOUGH INFO` outcomes and a decrease in `REFUTED` responses compared to NEFTune and LoRA. The combination of NEFTune and RAG + FD, however, exhibited a clear reduction in factual accuracy, marked by a \textsc{SUPPORTED} rate of 63.2\%. The sharply elevated \textsc{NOT ENOUGH INFO} and \textsc{REFUTED} rates may reflect the inability of the finetuned LLM to integrate additional text from the RAG process. The low RAG (raw) outcomes underscored the challenges inherent in leveraging unmodified RAG approaches and basic methodologies in achieving accurate retrieval for the purpose of synthesis of SLR-specific information that was generated during the data extraction process.

\begin{table}[h]
\centering
\fontsize{8pt}{10pt}\selectfont
\begin{tabular}{lccc}
\toprule
Method & REFUTED  & NOT ENOUGH INFO & SUPPORTED  \\ \hline
\midrule
NEFTune                     & 6.9\%  & 3.9\%   & 89.2\% \\
LoRA                        & 7.9\%  & 4.4\%   & 87.7\% \\
RAG + FD    & 4.6\%  & 10.7\%  & 84.7\% \\
NEFTune + (RAG + FD) & 10.4\% & 26.4 \%  & 63.2\% \\
RAG (raw)                   & 29.7\% & 52.0\%  & 18.3\% \\
Baseline &  25.8\% & 59.7\%  & 14.5\% \\
\bottomrule
\end{tabular}
\caption{Description of the FEVER results in rank-order according to the 'SUPPORTED' values shown as the percentage of responses according to each category are listed. }
\label{tab:fever}
\end{table}

The CGS results, detailed in Table~\ref{tab:cgs}, offer a more fine-grained perspective on the consistency of the responses. This scale ranges from -2 (Contradictory) to 2 (Fully consistent), providing a more detailed view of the alignment between the generated responses and the SLR corpus.
Similar to the FEVER results, in the CGS analysis NEFTune again emerged as the leader with the highest percentage of 'Fully consistent' responses and an overall mean score of 1.75 (87.6\%), indicating a predominance in the factual fidelity of the responses. LoRA closely followed with an overall mean score of 1.72 (86.0\%).   The method combining RAG with a fine-tuned dataset (RAG + FD) achieved a moderately lower mean score of 1.67 (83.7\%), reflecting a competent performance but with room for possible improvement via the adjustable RAG parameters, particularly in enhancing the proportion of `Fully consistent` responses. NEFTune combined with RAG + FD exhibited a notable drop in performance, as evidenced by a mean score of 1.14 (57.2\%), which suggests that the integration of RAG inputs with the finetuned models was likely introducing confusion in response generation. This significant decrease highlights the challenges in combining these two otherwise effective methodologies without compromising factual integrity. RAG (raw) and the baseline methods manifested the lowest consistency levels in line with previous results, with mean scores of 0.46 (23.2\%) and 0.164 (8.2\%) respectively.

\begin{table}[h]
\small
\centering
\begin{tabular}{lccccc|c}
\toprule
 &  & Partially   & Not enough   & Partially  & Fully  &  \\
  & Contradictory &  contradictory  &  info  &  consistent &  consistent &  \\
    & (-2) &  (-1)  &  (0)  &  (1) &  (2) & Mean \\ \hline
\midrule
NEFTune & 18 (0.4\%)& 205 (4.1\%) & 57 (1.1\%) & 432 (8.7\%) & 4250 (85.7\%) & 1.75 (87.6\%) \\
LoRA & 25 (0.5\%)& 215 (4.3\%) & 77 (1.6\%) & 486 (9.8\%) & 4159 (83.8\%) & 1.72 (86.0\%) \\
RAG + FD & 6 (0.1\%) & 86 (1.7\%) & 178 (3.6\%) & 983 (19.8\%) & 3709 (74.7\%) & 1.67 (83.7\%) \\
NEFTune + (RAG + FD) & 69 (1.4\%) & 250 (5.0\%) & 931 (18.8\%) & 1356 (27.3\%) & 2356 (47.5\%) & 1.14 (57.2\%) \\
RAG (raw) & 84 (1.7\%) & 713 (14.4\%) & 1286 (25.9\%) & 2579 (52.0\%) & 300 (6.0\%) & 0.46 (23.2\%) \\
Baseline & 341 (6.9\%) & 1001 (20.2\%) & 1397 (28.2\%) & 1948 (39.3\%) & 275 (5.5\%) & 0.164 (8.2\%) \\
\bottomrule
\end{tabular}
\caption{Description of the CGS results in rank-order according to the 'Fully consistent' values. The absolute number and the percentage of responses according to each category are listed. The numeric values are aggregated and listed as means ranging from 2 (perfectly correct) to -2 (entirely incorrect) as well as the percentage with respect to the highest possible score.}
\label{tab:cgs}
\end{table}

\subsection{Correlation analysis between responses of all approaches}

The correlation analysis between the responses of various LLM methodologies is presented next in order to explore similarities and divergences in the LLM responses that may offer insights complementary nature of each where there is evidence of them capturing different aspects of the data, leading to the potential of combining them to achieve better accuracies. The correlation matrix represented in Table~\ref{tab:fevercorrelation} offers a refined perspective on the ordinal associations between different LLM FEVER responses.

A positive \textit{k} value between NEFTune and LoRA indicates a moderately similar ranking order in their response evaluations. Conversely, a near-zero \textit{k} value between NEFTune and the baseline method suggests an independent performance trend, underscoring NEFTune's distinct methodological improvements over the benchmark. The correlations between responses of all other methods were not significant.

\begin{table}[hbt]
\small
\centering
\begin{tabular}{rcccccc}
\toprule
 & \text{NEFTune} & \text{RAG (raw) } & \text{LoRA} & \text{RAG + FD} & \text{NEFTune + (RAG + FD)} & \text{baseline} \\
\midrule
\text{NEFTune} & 1 &  &  &  &  \\
\text{RAG (raw) } & 0.013 & 1 &  &  &  \\
\text{LoRA} & 0.290 & 0.014 & 1 &  &  \\
\text{RAG + FD} & 0.128 & 0.037 & 0.116 & 1 &  \\
\text{NEFTune + (RAG + FD)} & 0.100 & 0.013 & 0.095 & 0.206 & 1 \\
\text{baseline} & -0.006 & 0.107 & 0.022 & 0.037 & -0.004 & 1 \\
\bottomrule
\end{tabular}
\caption{Correlation matrix for FEVER responses between all the methods.}
\label{tab:fevercorrelation}
\end{table}

Similar to the previous table, the CGS correlation Table \ref{tab:CGScorrelation}  again showed a moderately positive correlation between NEFTune and LoRA (0.342), indicating that these methods share exhibited overlapping learning patterns on the underlying finetuning dataset. The near-zero correlation between NEFTune and the baseline (-0.004) reinforces the distinct advancement NEFTune contributes. This independence from the baseline method and other approaches highlights NEFTune’s effectiveness in effectively encoding the finetuning dataset SLR.

\begin{table}[hbt]
\small
\centering
\begin{tabular}{rcccccc}
\toprule
 & \text{NEFTune} & \text{RAG (raw)} & \text{LoRA} & \text{RAG + FD} & \text{NEFTune + (RAG + FD)} & \text{baseline} \\
 \midrule
\text{NEFTune} & 1 &  &  &  &    \\
\text{RAG (raw)} & -0.012 & 1 &  &  &    \\
\text{LoRA} & 0.342 & 0.001 & 1 &  &    \\
\text{RAG + FD} & 0.129 & 0.084 & 0.137 & 1 &    \\
\text{NEFTune + (RAG + FD)} & 0.087 & 0.035 & 0.091 & 0.248 &  1  \\
\text{baseline} & -0.004 & 0.168 & 0.025 & 0.040 & 0.039  & 1 \\
\bottomrule
\end{tabular}
\caption{SCG correlation. }
\label{tab:CGScorrelation}
\end{table}

We next examine and compare responses of the top three best-performing approaches on two sample test questions from FEVER, showcasing an example where each method's response was classified as \textsc{NOT ENOUGH INFO} and \textsc{REFUTED}. With respect to the \textsc{NOT ENOUGH INFO} responses in Table \ref{tab:FEVERNoTEnough}, we find variances in detail and emphasis. In response to the question asking what manner does the dashboard contribute to the advancement of virtual learning, the reference text highlights the dashboard's role in managing lab activity schedules and resource usage on a shared AWS account, focusing on logistical aspects. In contrast, LoRA and NEFTune responses pivot towards the dashboard's educational support, emphasizing feedback, self-regulation, and progress monitoring, with NEFTune slightly touching on resource management but not to the extent mentioned in the reference. RAG + FD, while partially aligning with the reference by discussing resource usage management, also diverged by emphasizing feedback and activity completion, omitting the specific detail about scheduling. Thus, none of the responses fully encapsulated the reference's emphasis on time slot management and resource usage, illustrating a shift from logistical to educational functionalities in their interpretations. In these instances the classification of NOT ENOUGH INFO for all responses was correct.

{\fontsize{8pt}{10pt}\selectfont
\begin{longtable}{P{2cm}P{14cm}}
\hline
\textbf{} & \textbf{Text} \\
\hline
Question & According to the 2023SLR dataset, in the naranjo2019visualdashboard paper, in what manner does the dashboard contribute to the advancement of virtual learning?
 \\
Reference text & In the data used for the 2023SLR, for online instruction, the dashboard helps in identifying preferred time slots for students to engage in lab activities, given the self-paced nature of the instruction. This can aid in managing and preventing soft limits being reached due to concurrent resource usage in a shared AWS account by multiple students. Source: naranjo2019visualdashboard
 \\
 NEFTune & In the data used for the 2023SLR, the dashboard contributes to the advancement of online education by providing a visual tool that helps teachers and students monitor the progress of lab activities and the usage of resources in a hands-on learning approach. It attends specifically to learning metrics, supporting teacher interventions and fostering self-regulation among students. The dashboard also aids in scaling education by offering insights into student engagement and progress, potentially enabling the automation of evaluation processes and enhancing the learning experience. Source: naranjo2019visualdashboard
 \\
LoRA & In the data used for the 2023SLR, the dashboard contributes to virtual learning by providing timely and targeted feedback to students on their work, which supports self-regulation and learning about their own learning process. The dashboard's visual representation of learning activities and progress helps to engage students and increase their motivation. Additionally, by offering teachers a centralized platform for monitoring student progress, the dashboard facilitates timely intervention and personalized support. Source: naranjo2019visualdashboard
 \\
 RAG + FD & The dashboard contributes to the advancement of virtual learning by providing timely and targeted feedback on student work, allowing students to discriminate between completed parts of the hands-on labs and those still pending, fostering self-regulation and completion of activities. It also helps in managing and preventing soft limits being reached due to concurrent resource usage in a shared AWS account by multiple students.
 \\
 \bottomrule
 \caption{Example of responses 'NOT ENOUGH INFO' FEVER responses across top three methods. }
\label{tab:FEVERNoTEnough}
\end{longtable}
}

Meanwhile, with respect to the examples of \textsc{REFUTED} responses for the same question which requested details of a specific study's insights on how the learning analytics dashboard (LAD) drove the motivation among students in Table \ref{tab:FEVERRefuted}, we see that LoRA and NEFTune produced identical responses and closely match the reference text. However, while they correctly reported most details, they omitted the detail that intrinsic motivation actually decreased in both groups, as noted in the reference text. RAG + FD introduced a claim not present in the reference text, suggesting that LADs lead to a higher academic performance evidenced by higher final grades. Similar to LoRA and NEFTune, the response mentioned the increased extrinsic motivation in the treatment group and the lack of significant difference in changes in intrinsic motivation between the groups. However, it also omitted that intrinsic motivation decreased in both groups over time. Again, in these instances the FEVER classification of REFUTED derived by GPT-4 was correct for all responses.

{\fontsize{8pt}{10pt}\selectfont
\begin{longtable}{P{2cm}P{14cm}}
\hline
\textbf{} & \textbf{Text} \\
\hline
Question & According to the 2023SLR dataset, in the fleur2020motivation paper, could you detail the study's insights on how the LAD drives motivation among students?
 \\
Reference text & In the data used for the 2023SLR, the study found a significant interaction effect between group (treatment/control) and time (begin/end of the course) for extrinsic motivation, with the treatment group showing increased extrinsic motivation at the end of the course, while it decreased in the control group. However, intrinsic motivation decreased in both groups over time, with no significant difference between the groups. Source: fleur2020motivation
 \\
 NEFTune & In the data used for the 2023SLR, the study found a significant interaction effect between group (treatment/control) and time (beginning/end of the course) for extrinsic motivation, with the treatment group showing increased extrinsic motivation at the end of the course compared to the control group. However, there was no significant difference between the groups in terms of changes in intrinsic motivation. Source: fleur2020motivation
 \\
LoRA & In the data used for the 2023SLR, the study found a significant interaction effect between group (treatment/control) and time (beginning/end of the course) for extrinsic motivation, with the treatment group showing increased extrinsic motivation at the end of the course compared to the control group. However, there was no significant difference between the groups in terms of changes in intrinsic motivation. Source: fleur2020motivation
 \\
 RAG + FD & The study by Fleur et al. (2020) found that LADs can significantly increase extrinsic motivation among students, leading to higher academic performance. The treatment group showed increased extrinsic motivation at the end of the course compared to the control group, as evidenced by higher final grades. However, there was no significant difference between the groups in terms of changes in intrinsic motivation.
 \\
 \bottomrule
 \caption{Example of responses \textsc{REFUTED} FEVER responses across top three methods. }
\label{tab:FEVERRefuted}
\end{longtable}
}

\subsection{Detailed Analysis}

For the best-performing model from the above results, Tables \ref{tab:levels_summary} and \ref{tab:neftuneanalysis} offer a vantage point to assess NEFTune model's proficiency across various levels of analysis. The NEFTune model's capability is particularly observable at the paper-level Q\&As, with 83.0\% of responses achieving 'Fully Consistent' ratings (CGS=2) as shown in Table \ref{tab:levels_summary}. Likewise, at the paper-summary-level, NEFTune ability to respond with high fidelity is further confirmed by a 'Fully Consistent' score in 95.0\% of cases. When combined with 'Partially consistent' results, NEFTune's scores rise to 93\% and 98\% accuracy for paper-level and paper-summary-level respectively. However, for SLR-level questions, NEFTune's 'Fully Consistent' does significantly reduce to a score of 52\%, but again, in combination with 'Partially consistent' results, this rises to 84\%, still indicating a strong level of dependability.

\begin{table}[h]
\fontsize{8pt}{10pt}\selectfont
\centering

\begin{tabular}{rccccc}
\toprule
  & & Partially   & Not enough   & Partially  & Fully  \\
  & Contradictory &  contradictory  &  info  &  consistent &  consistent  \\
Level of analysis    & (-2) &  (-1)  &  (0)  &  (1) &  (2)  \\
\midrule
Paper-level & 11 (0.3\%) & 171 (4.7\%) & 55 (1.5\%) & 378 (10.4\%) & 3003 (83.0\%) \\
Paper-summary-level & 6 (0.5\%) & 23 (1.8\%) & 2 (0.2\%) & 33 (2.6\%) & 1211 (95.0\%) \\
SLR-level & 1 (1.3\%) & 11 (14.7\%) & - & 24 (32.0\%) & 39 (52.0\%) \\
\bottomrule
\end{tabular}
\caption{Summary of Levels by SCG for NEFTune}
\label{tab:levels_summary}
\end{table}

\begin{table}[h]
\fontsize{8pt}{10pt}\selectfont
\centering
\begin{tabular}{rccc}
\toprule
Level  of analysis & REFUTED & NOT ENOUGH INFO & SUPPORTED \\
\midrule
Paper-level &	179 (7.4\%) &	267 (4.9\%) &	3180 (87.7\%) \\
Paper-summary-level & 44 (3.5\%) & 14 (1.1\%) & 1217 (95.5\%) \\
SLR-level & 32 (42.7\%) & 1 (1.3\%) & 42 (56.0\%) \\
\bottomrule
\end{tabular}
\caption{Description of FEVER results at different levels of questions for NEFTune}
\label{tab:neftuneanalysis}
\end{table}

In the FEVER metric results in Table \ref{tab:neftuneanalysis}, the results do harmonize with the CGS findings, demonstrating the model's robust accuracy with a majority of 'Supported' classifications at both the paper level (87.7\%) and paper-summary level (95.5\%). Again, a divergent scenario unfolds at the SLR-level, where a considerable 42.7\% of responses are categorized as 'Refuted'.

The analyses of CGS and FEVER metrics, taken conjointly, paint a comprehensive overview of NEFTune's strengths. While the model's performance is robust in factual replication at paper-level Q\&As, the performance on SLR-level data points suggests an area for improvement in either the calibration of the FEVER metric, or in optimizing the learning for these data points by increasing the  size of the finetuning dataset for these samples. However, it should be noted that despite the lower accuracies and a reduced level of confidence in the model's outputs on the SLR-level questions, the SLR-level questions can nonetheless be verified and accurately reconstructed from more accurate paper-summary-level responses.

\subsection{Qualitative analysis accuracies}

The qualitative evaluations conducted manually across the eleven NEFTune LLM responses to questions about key SLR findings in the target paper representing the ground truth, reveal a high level of fidelity and alignment between the reference and the LLM's responses. An example of a typical response can be seen in Table \ref{tab:qualitativeanalysis1}. The analysis showed that in eight out of eleven cases, the key findings from the SLR ground truth could be faithfully extracted and reconstructed from the responses provided by the finetuned LLM's responses. However, in two instances, a deeper understanding of certain domain-specific concepts such as prescriptive analytics and model interpretability within the context of learning analytics was crucial to be able to arrive at ground truth conclusions from the LLM responses. These instances highlighted a divergence in the conceptualization of certain technical terms between the LLM and the SLR study, necessitating domain expertise to align the LLM's broader interpretations with the SLR's findings. Nonetheless, it can be concluded that in 90\% of the cases, the ground truth findings from the SLR study are supported by the LLM's responses. However, there was one notable LLM response that introduced some inaccuracies in its response concerning the number of participants across all the studies in 17 papers. It was only in his response that a deviation was observed and in this instance, it could be confirmed that the LLM response did not support the ground truth findings in the SLR study. Example responses from the qualitative analysis can be seen in Appendix \ref{qualitativeresponses}.

\begin{table}[h]
\centering
{\fontsize{8pt}{10pt}\selectfont
\begin{tabular}{p{0.1\linewidth}p{0.85\linewidth}}
\hline
\textbf{Qualitative evaluation} & \textbf{Response and analysis} \\
\hline
Question & According to the 2023SLR dataset, in the 2023SLR papers, were the reported dashboards (LADs) developed mostly in a pilot study state, or a prototype for academic purposes,  or were they placed into production? \\
\midrule
Ground truth & "Many of the published dashboards (LADs) are only in their prototype phases,and only few in the pilot implementation stages."  \\
\midrule
LLM & "In the data used for the 2023SLR, the majority of papers discuss the development and evaluation of LADs as a research exercise, either piloting the effectiveness of the dashboards or using them as a case study to investigate their utility. Papers that do not specifically report on the development of LADs as a research effort include Aljohani et al. (2019), Bodily et al. (2018), Karaoglanyilmaz et al. (2021), Fleur et al. (2020), Salehiankia et al. (2020), Han et al. (2020), Valle et al. (2021), Chatti et al. (2020), He et al. (2019), and Naranjo et al. (2019). Other papers, including Ulfa et al. (2019), Liu et al. (2019), Kokoç et al. (2021), Valle et al. (2021), and Baneres et al. (2019), provide details on the development and implementation of LADs, with some studying specific dashboards and others discussing the design process. Additionally, Owatari et al. (2020) and Aljohani et al. (2019) report on the development of LADs as advanced informational systems specifically designed for learning support. Source: 2023SLR"  \\
\midrule
Analysis & LLM's answer can be seen as an expanded version of the ground truth, maintaining the core assertion while adding context and detail. Therefore, the LLM’s response can be useed to reconstruct the ground truth for the SLR with a high degree of completeness and accuracy as it includes all the essential information of the ground truth, while extending it with specific examples and a broader examination of the LADs' development stages.  \\
\bottomrule
\end{tabular}
\caption{Comparison of Question and Ground Truth with LLM Response}
\label{tab:qualitativeanalysis1}
}
\end{table}

The analyses consistently demonstrate that the finetuned LLM's responses tend to provide a broader and more detailed perspective than ground truth which is often more concise and sometimes more definitive in its assertions. Below is a synthesis of the overarching patterns observed:

\begin{enumerate}
    \item Expansion and Detailing: The LLM frequently expands upon the information provided in ground truth introducing additional context, examples, and details that were not explicitly mentioned in the ground truth. This pattern is evident in analyses where the LLM's responses offer a deeper dive into specific studies, technologies, and outcomes associated with LADs, enhancing the understanding of the subject matter beyond the baseline established by ground truth. However, the additional information and details are correct and do not indicate hallucination.

    \item Alignment with Core Assertions: Despite the additional details, the LLM's responses generally align with the core assertions or findings of ground truth. This alignment indicates that the finetuned LLM successfully captures the essence of the ground truth findings, while adding value through elaboration and exemplification.

    \item Discrepancies in Emphasis and Scope: Some discrepancies arise primarily from differences in emphasis and scope. For instance, while the ground truth response might highlight a lack or absence of certain features or trends within the LADs research, the LLM occasionally points out exceptions or minor trends that counteract these broad strokes. These discrepancies do not necessarily contradict ground truth but rather suggest a more variegated picture, which can give the researcher more scope for expansion in an academic paper.

    \item Reconstruction and Completeness: The degree to which the ground truth can be reconstructed from the LLM's responses is generally very high, reflecting the LLM's ability to both mirror and in many responses, extend the ground truth in the target paper. In cases where the LLM's detailed accounts align closely with ground truth, the reconstruction ability is high, indicating a strong corroboration of ground truth by the LLM.

    \end{enumerate}

\section{Discussion}

This study has marked a pivotal shift towards advancing the use of AI and LLMs for automating systematic literature review methodologies, with a particular focus on the knowledge and evidence synthesis phase. The LLM finetuning approaches and the novel data extraction processes make a novel contribution to academic research practices, with even broader applications to other domains that require syntheses across multiple documents with the requirement of supporting question and answering tasks. This study has demonstrated that indeed modern PEFT finetuning approaches like NEFTune and LoRA can be used for finetuning effective LLMs (RQ1), and the proposed automated approach for extracting finetuning datasets from a corpus of selected academic papers can successfully construct finetuning datasets (RQ2) that support question and answer downstream tasks necessary for executing the final stages of an SLR. The study also affirms that the LLMs can be effectively finetuned on relatively small datasets (RQ3) that represent relatively narrow domains associated with the focused objectives of SLR studies.
A significant set of challenges that this study addressed and provided solutions for, concerned the broader problem of LLM hallucination and ensuring that the LLM's responses were solely based on the corpus of the selected academic papers, and how response provenance and the factuality of the responses could be evaluated (RQ4). We devised a novel token-based approach that enabled us to audit and verify the source information of the LLM responses. Meanwhile, we devised evaluation metrics for factuality and demonstrated how they can be effectively automated with the assistance of commercial LLMs for efficiency. Finally, the entire proposed framework for automating the final stages of an SLR was validated for viability by replicating an actual published SLR study, using it as a benchmark and source of ground truth. Our experiments indeed confirm the reliability of the proposed finetuning framework for its fidelity in replicating findings from an SLR study (RQ5).

The contributions from this study fill both an existing gap in the literature and a real need in academic research for automating SLR processes during the knowledge synthesis phase since most efforts have focused on using AI and machine learning to solve tasks such as paper searchers and filtering. While an increasing number of AI technologies are emerging to support academic research in the form of features enabling researchers to  \enquote{talk to} individual papers, they often have limitations and one of the key limitations is the inability to \enquote{talk across} multiple papers with high reliability and the ability to track the sources of all responses. These limitations and the lack of control over the existing commercial solutions raise issues of trust as well as replicability, which this study overcomes.

The solutions demonstrated in this study are not without their imperfections, but future developments can overcome them. The first involves the low inter-rater consonance achieved in this study, particularly for the FEVER metric which compromised some of its validity. This will be addressed in subsequent studies through a thorough training protocol that teaches the human evaluators with more examples of how to rate responses. The data extraction process can also be improved by extracting more than just a single Q\&A pair per chunk and section of each paper. This can instead be expanded to extracting multiple Q\&A pairs which will have the effect of encoding more details and information from each paper, and thus have a greater chance of capturing all the important facts from each study. The quality and capability of the finetuned models is to a significant degree a function of the data extraction phase. The LLM model can only know what it has been taught. Deficiencies and omissions in the data extraction step will hinder the depth of knowledge synthesis and findings that can be supported for an SLR, therefore,
most gains in the quality and factuality of LLM responses stand to be gained through the improvement of this process. Likewise, the lower accuracies exhibited on the SLR-level responses can also be improved in this manner by increasing the size of this category of Q\&A pairs.

Finally, it is inevitable with the current advancements in generative AI, and LLMs specifically, that these technologies are destined to increase their role in supporting and assisting in future SLR endeavors. To that end, there is a need for revisiting and updating the PRISMA reporting guidelines. This needs to be undertaken  in such a way that researchers in the future are provided with guidance on what needs to be reported when conducting PRISMA-conforming SLRs using AI automation tools in order to guarantee transparency and reproducibility. Such efforts have already begun \cite{susnjak2023prisma} and need to be finalised.

\subsection*{Study Limitations and Future Work}

This study's exploration of LLMs in SLRs was constrained by model selection, focusing on Mistral without assessing larger models over 7 billion parameters that might offer improved outcomes. Future research should broaden its scope to include a variety of open-source LLMs, evaluating their efficacy in SLR contexts.
Dataset access presents a significant limitation, as copyright restrictions by publishing houses impede the public sharing of even summarised content, challenging the reproducibility and validation of our findings. Addressing this issue requires navigating copyright laws to facilitate wider academic verification and exploration.
The use of GPT-4 for evaluation reflects the current state of technology, yet the rapid advancement in LLMs suggests that future iterations will necessitate updated prompts and methodologies to maintain relevance and accuracy in assessments. A limitation of using proprietary models like GPT-4 is also cost, which can and should be bypassed in the future using locally running LLMs for data extraction and evaluation.
The study’s focus on singular-topic questions limits the depth of evaluation regarding the models' ability to handle complex queries involving multiple concepts. Enhancing the CGS and FEVER metrics calibration through detailed training could yield more reliable evaluations, a necessity for future studies to consider.
Methodologically, the study leaned towards qualitative meta-synthesis, leaving a gap in quantitative meta-analysis exploration. Addressing this gap, alongside refining hyperparameters and expanding computational resources to better support extensive data processing, forms a critical pathway for subsequent research. Performing hyperparameter tuning was particularly challenging given the long training runtimes for model finetuning, and it is highly likely that suboptimal hyperparameters were used in this research as a result.
Future work should encompass large-scale replication studies to confirm and extend these initial findings, incorporating both qualitative and quantitative SLR meta-analyses. This includes investigating LLMs' multimodal capabilities in extracting information from diverse formats such as tables and figures.

\section{Conclusion}

This research introduces an SLR-automation framework leveraging finetuned LLMs, presenting a novel and significant methodical advancement in employing AI for academic research. Our comprehensive experiments with LLM finetuning demonstrate that these AI technologies can effectively streamline SLRs, ensuring both efficiency and accuracy in information retrieval. The framework's effectiveness was validated by accurately replicating a pre-existing SLR study, showcasing the practical applicability of our methods.

The study not only underscores the potential of AI to enhance the SLR process through systematic and efficient literature synthesis but also addresses critical challenges such as LLM hallucination and data provenance. By tackling these issues, we ensure the reliability and verifiability of the synthesized content.

In proposing this SLR-automation framework, we contribute to the broader discourse on integrating AI-driven methodologies in academic research, advocating for the update of PRISMA guidelines to encapsulate these advanced techniques. This ensures methodological transparency and rigor in future SLRs. Our work lays a foundation for further exploration in this area, highlighting the necessity for continuous development of AI tools to enrich and facilitate scholarly research.

\bibliographystyle{unsrtnat}

\begin{thebibliography}{82}
\providecommand{\natexlab}[1]{#1}
\providecommand{\url}[1]{\texttt{#1}}
\expandafter\ifx\csname urlstyle\endcsname\relax
  \providecommand{\doi}[1]{doi: #1}\else
  \providecommand{\doi}{doi: \begingroup \urlstyle{rm}\Url}\fi

\bibitem[Paez(2017)]{paez2017Grey}
Arsenio Paez.
\newblock Grey literature: An important resource in systematic reviews.
\newblock \emph{Journal of evidence-based medicine}, 2017.
\newblock \doi{10.1111/jebm.12265}.

\bibitem[Xiao and Watson(2017)]{Xiao2017Guidance}
Yu~Xiao and M.~Watson.
\newblock Guidance on conducting a systematic literature review.
\newblock \emph{Journal of Planning Education and Research}, 39:\penalty0 112 -- 93, 2017.
\newblock \doi{10.1177/0739456X17723971}.

\bibitem[Siddaway et~al.(2019)Siddaway, Wood, and Hedges]{Siddaway2019How}
Andy~P Siddaway, A.~Wood, and L.~Hedges.
\newblock How to do a systematic review: A best practice guide for conducting and reporting narrative reviews, meta-analyses, and meta-syntheses.
\newblock \emph{Annual review of psychology}, 70:\penalty0 747--770, 2019.
\newblock \doi{10.1146/annurev-psych-010418-102803}.

\bibitem[Page et~al.(2021)Page, McKenzie, Bossuyt, Boutron, Hoffmann, Mulrow, Shamseer, Tetzlaff, Akl, Brennan, et~al.]{page2021prisma}
Matthew~J Page, Joanne~E McKenzie, Patrick~M Bossuyt, Isabelle Boutron, Tammy~C Hoffmann, Cynthia~D Mulrow, Larissa Shamseer, Jennifer~M Tetzlaff, Elie~A Akl, Sue~E Brennan, et~al.
\newblock The prisma 2020 statement: an updated guideline for reporting systematic reviews.
\newblock \emph{International journal of surgery}, 88:\penalty0 105906, 2021.

\bibitem[Liberati et~al.(2009)Liberati, Altman, Tetzlaff, Mulrow, Gøtzsche, Ioannidis, Clarke, Clarke, Devereaux, Kleijnen, and Moher]{Liberati2009The}
A.~Liberati, D.~Altman, J.~Tetzlaff, C.~Mulrow, P.~Gøtzsche, J.~Ioannidis, M.~Clarke, M.~Clarke, P.~J. Devereaux, J.~Kleijnen, and D.~Moher.
\newblock The prisma statement for reporting systematic reviews and meta-analyses of studies that evaluate health care interventions: explanation and elaboration.
\newblock \emph{Journal of clinical epidemiology}, 62 10:\penalty0 e1--34, 2009.
\newblock \doi{10.1016/j.jclinepi.2009.06.006}.

\bibitem[Rethlefsen et~al.(2021)Rethlefsen, Kirtley, Waffenschmidt, Ayala, Moher, Page, and Koffel]{Rethlefsen2021PRISMAS}
M.~Rethlefsen, S.~Kirtley, S.~Waffenschmidt, A.~P. Ayala, D.~Moher, M.~Page, and Jonathan~B. Koffel.
\newblock Prisma-s: an extension to the prisma statement for reporting literature searches in systematic reviews.
\newblock \emph{Journal of the Medical Library Association : JMLA}, 109:\penalty0 174 -- 200, 2021.
\newblock \doi{10.5195/jmla.2021.962}.

\bibitem[Williams et~al.(2020)Williams, Clark, Clark, and Raffo]{Williams2020Reexamining}
Ralph~I. Williams, L.~Clark, W.~R. Clark, and Deana~M. Raffo.
\newblock Re-examining systematic literature review in management research: Additional benefits and execution protocols.
\newblock \emph{European Management Journal}, 2020.
\newblock \doi{10.1016/J.EMJ.2020.09.007}.

\bibitem[de~la Torre-L{\'o}pez et~al.(2023)de~la Torre-L{\'o}pez, Ram{\'\i}rez, and Romero]{de2023artificial}
Jos{\'e} de~la Torre-L{\'o}pez, Aurora Ram{\'\i}rez, and Jos{\'e}~Ra{\'u}l Romero.
\newblock Artificial intelligence to automate the systematic review of scientific literature.
\newblock \emph{Computing}, 105\penalty0 (10):\penalty0 2171--2194, 2023.

\bibitem[Chen and Song(2019)]{Chen2019Visualizing}
Chaomei Chen and Min Song.
\newblock Visualizing a field of research: A methodology of systematic scientometric reviews.
\newblock \emph{PLoS ONE}, 14, 2019.
\newblock \doi{10.1371/journal.pone.0223994}.

\bibitem[Ganann et~al.(2010)Ganann, Ciliska, and Thomas]{Ganann2010Expediting}
R.~Ganann, D.~Ciliska, and H.~Thomas.
\newblock Expediting systematic reviews: methods and implications of rapid reviews.
\newblock \emph{Implementation Science : IS}, 5:\penalty0 56 -- 56, 2010.
\newblock \doi{10.1186/1748-5908-5-56}.

\bibitem[McInnes et~al.(1999)McInnes, Duf, and McClarey]{McInnes1999Challenges}
E.~McInnes, Lesley Duf, and M.~McClarey.
\newblock Challenges in updating a systematic review.
\newblock \emph{Nursing Times Research}, 4:\penalty0 66 -- 71, 1999.
\newblock \doi{10.1177/136140969900400111}.

\bibitem[Mahood et~al.(2014)Mahood, Eerd, and Irvin]{Mahood2014Searching}
Quenby Mahood, D.~Van Eerd, and E.~Irvin.
\newblock Searching for grey literature for systematic reviews: challenges and benefits.
\newblock \emph{Research Synthesis Methods}, 5:\penalty0 221 -- 234, 2014.
\newblock \doi{10.1002/jrsm.1106}.

\bibitem[Elamin et~al.(2009)Elamin, Flynn, Bassler, Briel, Alonso-Coello, Karanicolas, Guyatt, Málaga, Furukawa, Kunz, Schuuenemann, Murad, Barbui, Cipriani, and Montori]{Elamin2009Choice}
Mohamed~B. Elamin, David~N. Flynn, D.~Bassler, M.~Briel, P.~Alonso-Coello, P.~Karanicolas, G.~Guyatt, G.~Málaga, T.~Furukawa, R.~Kunz, H.~Schuuenemann, M.~Murad, C.~Barbui, A.~Cipriani, and V.~Montori.
\newblock Choice of data extraction tools for systematic reviews depends on resources and review complexity.
\newblock \emph{Journal of clinical epidemiology}, 62 5:\penalty0 506--10, 2009.
\newblock \doi{10.1016/j.jclinepi.2008.10.016}.

\bibitem[Bui et~al.(2016)Bui, Del~Fiol, Hurdle, and Jonnalagadda]{bui2016extractive}
Duy Duc~An Bui, Guilherme Del~Fiol, John~F Hurdle, and Siddhartha Jonnalagadda.
\newblock Extractive text summarization system to aid data extraction from full text in systematic review development.
\newblock \emph{Journal of biomedical informatics}, 64:\penalty0 265--272, 2016.

\bibitem[Susnjak(2023)]{susnjak2023prisma}
Teo Susnjak.
\newblock Prisma-dfllm: An extension of prisma for systematic literature reviews using domain-specific finetuned large language models.
\newblock \emph{arXiv preprint arXiv:2306.14905}, 2023.

\bibitem[Min et~al.(2023)Min, Ross, Sulem, Veyseh, Nguyen, Sainz, Agirre, Heintz, and Roth]{bonan2023recent}
Bonan Min, Hayley Ross, Elior Sulem, Amir Pouran~Ben Veyseh, Thien~Huu Nguyen, Oscar Sainz, Eneko Agirre, Ilana Heintz, and Dan Roth.
\newblock Recent advances in natural language processing via large pre-trained language models: A survey.
\newblock \emph{ACM Comput. Surv.}, 56\penalty0 (2), sep 2023.
\newblock ISSN 0360-0300.
\newblock \doi{10.1145/3605943}.
\newblock URL \url{https://doi.org/10.1145/3605943}.

\bibitem[Chang et~al.(2024)Chang, Wang, Wang, Wu, Yang, Zhu, Chen, Yi, Wang, Wang, Ye, Zhang, Chang, Yu, Yang, and Xie]{Yupeng2024llmsurvey}
Yupeng Chang, Xu~Wang, Jindong Wang, Yuan Wu, Linyi Yang, Kaijie Zhu, Hao Chen, Xiaoyuan Yi, Cunxiang Wang, Yidong Wang, Wei Ye, Yue Zhang, Yi~Chang, Philip~S. Yu, Qiang Yang, and Xing Xie.
\newblock A survey on evaluation of large language models.
\newblock \emph{ACM Trans. Intell. Syst. Technol.}, 15\penalty0 (3), mar 2024.
\newblock ISSN 2157-6904.
\newblock \doi{10.1145/3641289}.
\newblock URL \url{https://doi.org/10.1145/3641289}.

\bibitem[Hou et~al.(2023)Hou, Zhao, Liu, Yang, Wang, Li, Luo, Lo, Grundy, and Wang]{Hou2023Large}
Xinying Hou, Yanjie Zhao, Yue Liu, Zhou Yang, Kailong Wang, Li~Li, Xiapu Luo, David Lo, John~C. Grundy, and Haoyu Wang.
\newblock Large language models for software engineering: A systematic literature review.
\newblock \emph{ArXiv}, abs/2308.10620, 2023.
\newblock \doi{10.48550/arXiv.2308.10620}.

\bibitem[Kalai and Vempala(2023)]{kalai2023calibrated}
Adam~Tauman Kalai and Santosh~S Vempala.
\newblock Calibrated language models must hallucinate.
\newblock \emph{arXiv preprint arXiv:2311.14648}, 2023.

\bibitem[McIntosh et~al.(2023{\natexlab{a}})McIntosh, Liu, Susnjak, Watters, Ng, and Halgamuge]{McIntosh2023hallucination}
Timothy~R. McIntosh, Tong Liu, Teo Susnjak, Paul Watters, Alex Ng, and Malka~N. Halgamuge.
\newblock A culturally sensitive test to evaluate nuanced gpt hallucination.
\newblock \emph{IEEE Transactions on Artificial Intelligence}, pages 1--13, 2023{\natexlab{a}}.
\newblock \doi{10.1109/TAI.2023.3332837}.

\bibitem[Li et~al.(2023)Li, Du, Zhou, Wang, Zhao, and rong Wen]{Li2023Evaluating}
Yifan Li, Yifan Du, Kun Zhou, Jinpeng Wang, Wayne~Xin Zhao, and Ji~rong Wen.
\newblock Evaluating object hallucination in large vision-language models.
\newblock \emph{ArXiv}, pages 292--305, 2023.
\newblock \doi{10.48550/arXiv.2305.10355}.

\bibitem[Nashwan and Jaradat(2023)]{Nashwan2023Streamlining}
A.~Nashwan and Jaber~H Jaradat.
\newblock Streamlining systematic reviews: Harnessing large language models for quality assessment and risk-of-bias evaluation.
\newblock \emph{Cureus}, 15, 2023.
\newblock \doi{10.7759/cureus.43023}.

\bibitem[Khraisha et~al.(2023)Khraisha, Put, Kappenberg, Warraitch, and Hadfield]{Khraisha2023Can}
Qusai Khraisha, Sophie Put, Johanna Kappenberg, Azza Warraitch, and Kristin Hadfield.
\newblock Can large language models replace humans in the systematic review process? evaluating gpt-4's efficacy in screening and extracting data from peer-reviewed and grey literature in multiple languages.
\newblock \emph{ArXiv}, abs/2310.17526, 2023.
\newblock \doi{10.48550/arXiv.2310.17526}.

\bibitem[Fernandez et~al.(2023)Fernandez, Elmore, Franklin, Krishnan, and Tan]{Fernandez2023How}
R.~Fernandez, Aaron~J. Elmore, M.~Franklin, Sanjay Krishnan, and Chenhao Tan.
\newblock How large language models will disrupt data management.
\newblock \emph{Proc. VLDB Endow.}, 16:\penalty0 3302--3309, 2023.
\newblock \doi{10.14778/3611479.3611527}.

\bibitem[Peng et~al.(2023)Peng, Rousseau, Shortliffe, and Weng]{peng2023ai}
Yifan Peng, Justin~F Rousseau, Edward~H Shortliffe, and Chunhua Weng.
\newblock Ai-generated text may have a role in evidence-based medicine.
\newblock \emph{Nature medicine}, 29\penalty0 (7):\penalty0 1593--1594, 2023.

\bibitem[Smith(2024)]{smith2024reviews}
Linda~C Smith.
\newblock Reviews and reviewing: Approaches to research synthesis. an annual review of information science and technology (arist) paper.
\newblock \emph{Journal of the Association for Information Science and Technology}, 75\penalty0 (3):\penalty0 245--267, 2024.

\bibitem[Qureshi et~al.(2023)Qureshi, Shaughnessy, Gill, Robinson, Li, and Agai]{qureshi2023chatgpt}
Riaz Qureshi, Daniel Shaughnessy, Kayden~AR Gill, Karen~A Robinson, Tianjing Li, and Eitan Agai.
\newblock Are chatgpt and large language models “the answer” to bringing us closer to systematic review automation?
\newblock \emph{Systematic Reviews}, 12\penalty0 (1):\penalty0 72, 2023.

\bibitem[Bolanos et~al.(2024)Bolanos, Salatino, Osborne, and Motta]{bolanos2024artificial}
Francisco Bolanos, Angelo Salatino, Francesco Osborne, and Enrico Motta.
\newblock Artificial intelligence for literature reviews: Opportunities and challenges.
\newblock \emph{arXiv preprint arXiv:2402.08565}, 2024.

\bibitem[Okoli(2015)]{okoli2015guide}
Chitu Okoli.
\newblock A guide to conducting a standalone systematic literature review.
\newblock \emph{Communications of the Association for Information Systems}, 37, 2015.

\bibitem[Grant and Booth(2009)]{grant2009typology}
Maria~J Grant and Andrew Booth.
\newblock A typology of reviews: an analysis of 14 review types and associated methodologies.
\newblock \emph{Health information \& libraries journal}, 26\penalty0 (2):\penalty0 91--108, 2009.

\bibitem[Torraco(2016)]{torraco2016writing}
Richard~J Torraco.
\newblock Writing integrative literature reviews: Using the past and present to explore the future.
\newblock \emph{Human resource development review}, 15\penalty0 (4):\penalty0 404--428, 2016.

\bibitem[Torraco(2005)]{torraco2005writing}
Richard~J Torraco.
\newblock Writing integrative literature reviews: Guidelines and examples.
\newblock \emph{Human resource development review}, 4\penalty0 (3):\penalty0 356--367, 2005.

\bibitem[Dixon-Woods et~al.(2005)Dixon-Woods, Agarwal, Jones, Young, and Sutton]{dixon2005synthesising}
Mary Dixon-Woods, Shona Agarwal, David Jones, Bridget Young, and Alex Sutton.
\newblock Synthesising qualitative and quantitative evidence: a review of possible methods.
\newblock \emph{Journal of health services research \& policy}, 10\penalty0 (1):\penalty0 45--53, 2005.

\bibitem[Gurevitch et~al.(2018)Gurevitch, Koricheva, Nakagawa, and Stewart]{Gurevitch2018Metaanalysis}
J.~Gurevitch, J.~Koricheva, Shinichi Nakagawa, and G.~Stewart.
\newblock Meta-analysis and the science of research synthesis.
\newblock \emph{Nature}, 555:\penalty0 175--182, 2018.
\newblock \doi{10.1038/nature25753}.

\bibitem[Haddaway and Rytwinski(2018)]{Haddaway2018Metaanalysis}
Neal~R Haddaway and T.~Rytwinski.
\newblock Meta-analysis is not an exact science: Call for guidance on quantitative synthesis decisions.
\newblock \emph{Environment international}, 114:\penalty0 357--359, 2018.
\newblock \doi{10.1016/j.envint.2018.02.018}.

\bibitem[Metelli and Chaimani(2020)]{Metelli2020Challenges}
S.~Metelli and A.~Chaimani.
\newblock Challenges in meta-analyses with observational studies.
\newblock \emph{Evidence-Based Mental Health}, 23:\penalty0 83 -- 87, 2020.
\newblock \doi{10.1136/ebmental-2019-300129}.

\bibitem[Xu(2008)]{Xu2008Methodological}
Y.~Xu.
\newblock Methodological issues and challenges in data collection and analysis of qualitative meta-synthesis.
\newblock \emph{Asian nursing research}, 2 3:\penalty0 173--83, 2008.
\newblock \doi{10.1016/S1976-1317(08)60041-9}.

\bibitem[Mohammed et~al.(2016)Mohammed, Moles, and Chen]{Mohammed2016Metasynthesis}
Mohammed~A Mohammed, R.~Moles, and T.~Chen.
\newblock Meta-synthesis of qualitative research: the challenges and opportunities.
\newblock \emph{International Journal of Clinical Pharmacy}, 38:\penalty0 695--704, 2016.
\newblock \doi{10.1007/s11096-016-0289-2}.

\bibitem[Zimmer(2006)]{Zimmer2006Qualitative}
Lela~V. Zimmer.
\newblock Qualitative meta-synthesis: a question of dialoguing with texts.
\newblock \emph{Journal of advanced nursing}, 53 3:\penalty0 311--8, 2006.
\newblock \doi{10.1111/J.1365-2648.2006.03721.X}.

\bibitem[Hong et~al.(2017)Hong, Pluye, Bujold, and Wassef]{Hong2017Convergent}
Q.~Hong, P.~Pluye, Mathieu Bujold, and M.~Wassef.
\newblock Convergent and sequential synthesis designs: implications for conducting and reporting systematic reviews of qualitative and quantitative evidence.
\newblock \emph{Systematic Reviews}, 6, 2017.
\newblock \doi{10.1186/s13643-017-0454-2}.

\bibitem[Brunton et~al.(2020)Brunton, Oliver, and Thomas]{Brunton2020Innovations}
G.~Brunton, S.~Oliver, and James Thomas.
\newblock Innovations in framework synthesis as a systematic review method.
\newblock \emph{Research Synthesis Methods}, 11:\penalty0 316 -- 330, 2020.
\newblock \doi{10.1002/jrsm.1399}.

\bibitem[Flemming et~al.(2019)Flemming, Booth, Garside, Tunçalp, and Noyes]{Flemming2019Qualitative}
K.~Flemming, A.~Booth, R.~Garside, Ö. Tunçalp, and J.~Noyes.
\newblock Qualitative evidence synthesis for complex interventions and guideline development: clarification of the purpose, designs and relevant methods.
\newblock \emph{BMJ Global Health}, 4, 2019.
\newblock \doi{10.1136/bmjgh-2018-000882}.

\bibitem[Higgins et~al.(2019)Higgins, López-López, Becker, Davies, Dawson, Grimshaw, McGuinness, Moore, Rehfuess, Thomas, and Caldwell]{Higgins2019Synthesising}
J.~Higgins, J.~López-López, B.~Becker, S.~Davies, S.~Dawson, J.~Grimshaw, L.~McGuinness, T.~Moore, E.~Rehfuess, James Thomas, and D.~Caldwell.
\newblock Synthesising quantitative evidence in systematic reviews of complex health interventions.
\newblock \emph{BMJ Global Health}, 4, 2019.
\newblock \doi{10.1136/bmjgh-2018-000858}.

\bibitem[Jonnalagadda et~al.(2015)Jonnalagadda, Goyal, and Huffman]{Jonnalagadda2015}
Siddhartha~R Jonnalagadda, Pankaj Goyal, and Mark~D Huffman.
\newblock Automating data extraction in systematic reviews: a systematic review.
\newblock \emph{Systematic Reviews}, 4\penalty0 (1), 2015.

\bibitem[Van~Altena et~al.(2019)Van~Altena, Spijker, and Olabarriaga]{van2019usage}
AJ~Van~Altena, R~Spijker, and SD~Olabarriaga.
\newblock Usage of automation tools in systematic reviews.
\newblock \emph{Research synthesis methods}, 10\penalty0 (1):\penalty0 72--82, 2019.

\bibitem[Feng et~al.(2017)Feng, Chiam, and Lo]{feng2017text}
Luyi Feng, Yin~Kia Chiam, and Sin~Kuang Lo.
\newblock Text-mining techniques and tools for systematic literature reviews: A systematic literature review.
\newblock In \emph{2017 24th asia-pacific software engineering conference (apsec)}, pages 41--50. IEEE, 2017.

\bibitem[M{\"u}ller et~al.(2022)M{\"u}ller, Pachnanda, Pahl, and Rosenqvist]{muller2022application}
Henry M{\"u}ller, Simran Pachnanda, Felix Pahl, and Christopher Rosenqvist.
\newblock The application of artificial intelligence on different types of literature reviews-a comparative study.
\newblock In \emph{2022 International Conference on Applied Artificial Intelligence (ICAPAI)}, pages 1--7. IEEE, 2022.

\bibitem[van Dinter et~al.(2021)van Dinter, Tekinerdogan, and Catal]{van2021automation}
Raymon van Dinter, Bedir Tekinerdogan, and Cagatay Catal.
\newblock Automation of systematic literature reviews: A systematic literature review.
\newblock \emph{Information and Software Technology}, 136:\penalty0 106589, 2021.

\bibitem[Sundaram and Berleant(2023)]{sundaram2023automating}
Girish Sundaram and Daniel Berleant.
\newblock Automating systematic literature reviews with natural language processing and text mining: A systematic literature review.
\newblock In \emph{International Congress on Information and Communication Technology}, pages 73--92. Springer, 2023.

\bibitem[da~Silva~J{\'u}nior and Dutra(2021)]{da2021roadmap}
Eug{\^e}nio~Monteiro da~Silva~J{\'u}nior and Mois{\'e}s~Lima Dutra.
\newblock A roadmap toward the automatic composition of systematic literature reviews.
\newblock \emph{Iberoamerican Journal of Science Measurement and Communication}, 1\penalty0 (2):\penalty0 1--22, 2021.

\bibitem[Wagner et~al.(2022)Wagner, Lukyanenko, and Par{\'e}]{wagner2022artificial}
Gerit Wagner, Roman Lukyanenko, and Guy Par{\'e}.
\newblock Artificial intelligence and the conduct of literature reviews.
\newblock \emph{Journal of Information Technology}, 37\penalty0 (2):\penalty0 209--226, 2022.

\bibitem[Tsafnat et~al.(2014)Tsafnat, Glasziou, Choong, Dunn, Galgani, and Coiera]{tsafnat2014systematic}
Guy Tsafnat, Paul Glasziou, Miew~Keen Choong, Adam Dunn, Filippo Galgani, and Enrico Coiera.
\newblock Systematic review automation technologies.
\newblock \emph{Systematic reviews}, 3:\penalty0 1--15, 2014.

\bibitem[Marshall and Wallace(2019)]{marshall2019toward}
Iain~J Marshall and Byron~C Wallace.
\newblock Toward systematic review automation: a practical guide to using machine learning tools in research synthesis.
\newblock \emph{Systematic reviews}, 8:\penalty0 1--10, 2019.

\bibitem[Atkinson(2023)]{atkinson2023cheap}
Cameron~F Atkinson.
\newblock Cheap, quick, and rigorous: artificial intelligence and the systematic literature review.
\newblock \emph{Social Science Computer Review}, page 08944393231196281, 2023.

\bibitem[Altmami and Menai(2022)]{altmami2022automatic}
Nouf~Ibrahim Altmami and Mohamed El~Bachir Menai.
\newblock Automatic summarization of scientific articles: A survey.
\newblock \emph{Journal of King Saud University-Computer and Information Sciences}, 34\penalty0 (4):\penalty0 1011--1028, 2022.

\bibitem[Vaswani et~al.(2017)Vaswani, Shazeer, Parmar, Uszkoreit, Jones, Gomez, Kaiser, and Polosukhin]{vaswani2017attention}
Ashish Vaswani, Noam Shazeer, Niki Parmar, Jakob Uszkoreit, Llion Jones, Aidan~N Gomez, {\L}ukasz Kaiser, and Illia Polosukhin.
\newblock Attention is all you need.
\newblock \emph{Advances in neural information processing systems}, 30, 2017.

\bibitem[Brown et~al.(2020)Brown, Mann, Ryder, Subbiah, Kaplan, Dhariwal, Neelakantan, Shyam, Sastry, Askell, et~al.]{brown2020language}
Tom~B Brown, Benjamin Mann, Nick Ryder, Melanie Subbiah, Jared Kaplan, Prafulla Dhariwal, Arvind Neelakantan, Pranav Shyam, Girish Sastry, Amanda Askell, et~al.
\newblock Language models are few-shot learners.
\newblock \emph{arXiv preprint arXiv:2005.14165}, 2020.

\bibitem[Gupta et~al.(2023)Gupta, Herzog, Weisberger, Chao, Chaiyasate, and Lee]{gupta2023utilization}
Rohun Gupta, Isabel Herzog, Joseph Weisberger, John Chao, Kongkrit Chaiyasate, and Edward~S Lee.
\newblock Utilization of chatgpt for plastic surgery research: friend or foe?
\newblock \emph{Journal of Plastic, Reconstructive \& Aesthetic Surgery}, 80:\penalty0 145--147, 2023.

\bibitem[Hill et~al.(2023)Hill, Harris, and Clegg]{hill2023methods}
James~Edward Hill, Catherine Harris, and Andrew Clegg.
\newblock Methods for using bing's ai-powered search engine for data extraction for a systematic review.
\newblock \emph{Research Synthesis Methods}, 2023.

\bibitem[Castillo-Segura et~al.(2023)Castillo-Segura, Alario-Hoyos, Kloos, and Panadero]{castillo2023leveraging}
Pablo Castillo-Segura, Carlos Alario-Hoyos, Carlos~Delgado Kloos, and Carmen~Fern{\'a}ndez Panadero.
\newblock Leveraging the potential of generative ai to accelerate systematic literature reviews: An example in the area of educational technology.
\newblock In \emph{2023 World Engineering Education Forum-Global Engineering Deans Council (WEEF-GEDC)}, pages 1--8. IEEE, 2023.

\bibitem[Kumar(2023)]{kumar2023analysis}
Arun~HS Kumar.
\newblock Analysis of chatgpt tool to assess the potential of its utility for academic writing in biomedical domain.
\newblock \emph{Biology, Engineering, Medicine and Science Reports}, 9\penalty0 (1):\penalty0 24--30, 2023.

\bibitem[Zimmermann et~al.(2024)Zimmermann, Staab, Nasseri, and Brandtner]{zimmermann2024leveraging}
Robert Zimmermann, Marina Staab, Mehran Nasseri, and Patrick Brandtner.
\newblock Leveraging large language models for literature review tasks-a case study using chatgpt.
\newblock In \emph{International Conference on Advanced Research in Technologies, Information, Innovation and Sustainability}, pages 313--323. Springer, 2024.

\bibitem[Alshami et~al.(2023)Alshami, Elsayed, Ali, Eltoukhy, and Zayed]{alshami2023harnessing}
Ahmad Alshami, Moustafa Elsayed, Eslam Ali, Abdelrahman~EE Eltoukhy, and Tarek Zayed.
\newblock Harnessing the power of chatgpt for automating systematic review process: Methodology, case study, limitations, and future directions.
\newblock \emph{Systems}, 11\penalty0 (7):\penalty0 351, 2023.

\bibitem[Najafali et~al.(2023)Najafali, Camacho, Reiche, Galbraith, Morrison, and Dorafshar]{najafali2023truth}
Daniel Najafali, Justin~M Camacho, Erik Reiche, Logan~G Galbraith, Shane~D Morrison, and Amir~H Dorafshar.
\newblock Truth or lies? the pitfalls and limitations of chatgpt in systematic review creation.
\newblock \emph{Aesthetic Surgery Journal}, 43\penalty0 (8):\penalty0 NP654--NP655, 2023.

\bibitem[Lin et~al.(2022)Lin, Hilton, and Evans]{lin2022truthfulqa}
Stephanie Lin, Jacob Hilton, and Owain Evans.
\newblock {T}ruthful{QA}: Measuring how models mimic human falsehoods.
\newblock In Smaranda Muresan, Preslav Nakov, and Aline Villavicencio, editors, \emph{Proceedings of the 60th Annual Meeting of the Association for Computational Linguistics (Volume 1: Long Papers)}, pages 3214--3252, Dublin, Ireland, May 2022. Association for Computational Linguistics.
\newblock \doi{10.18653/v1/2022.acl-long.229}.
\newblock URL \url{https://aclanthology.org/2022.acl-long.229}.

\bibitem[Aksitov et~al.(2023)Aksitov, Chang, Reitter, Shakeri, and Sung]{aksitov2023characterizing}
Renat Aksitov, Chung-Ching Chang, David Reitter, Siamak Shakeri, and Yunhsuan Sung.
\newblock Characterizing attribution and fluency tradeoffs for retrieval-augmented large language models, 2023.

\bibitem[Zhang et~al.(2023)Zhang, Press, Merrill, Liu, and Smith]{zhang2023language}
Muru Zhang, Ofir Press, William Merrill, Alisa Liu, and Noah~A. Smith.
\newblock How language model hallucinations can snowball, 2023.

\bibitem[Lewis et~al.(2020)Lewis, Perez, Piktus, Petroni, Karpukhin, Goyal, Kuttler, Lewis, tau Yih, Rocktäschel, Riedel, and Kiela]{Lewis2020RetrievalAugmented}
Patrick Lewis, Ethan Perez, Aleksandara Piktus, Fabio Petroni, Vladimir Karpukhin, Naman Goyal, Heinrich Kuttler, M.~Lewis, Wen tau Yih, Tim Rocktäschel, Sebastian Riedel, and Douwe Kiela.
\newblock Retrieval-augmented generation for knowledge-intensive nlp tasks.
\newblock \emph{ArXiv}, abs/2005.11401, 2020.

\bibitem[Asai et~al.(2023)Asai, Wu, Wang, Sil, and Hajishirzi]{asai2023SelfRAG:}
Akari Asai, Zeqiu Wu, Yizhong Wang, Avirup Sil, and Hannaneh Hajishirzi.
\newblock Self-rag: Learning to retrieve, generate, and critique through self-reflection.
\newblock \emph{ArXiv}, abs/2310.11511, 2023.
\newblock \doi{10.48550/arXiv.2310.11511}.

\bibitem[Gururangan et~al.(2020)Gururangan, Marasović, Swayamdipta, Lo, Beltagy, Downey, and Smith]{Gururangan2020}
Suchin Gururangan, Ana Marasović, Swabha Swayamdipta, Kyle Lo, Iz~Beltagy, Doug Downey, and Noah~A Smith.
\newblock Don’t stop pretraining: Adapt language models to domains and tasks.
\newblock \emph{arXiv preprint arXiv:2004.10964}, 2020.

\bibitem[Howard and Ruder(2018)]{howard2018universal}
Jeremy Howard and Sebastian Ruder.
\newblock Universal language model fine-tuning for text classification.
\newblock \emph{arXiv preprint arXiv:1801.06146}, 2018.

\bibitem[Houlsby et~al.(2019)Houlsby, Giurgiu, Jastrzebski, Morrone, De~Laroussilhe, Gesmundo, Attariyan, and Gelly]{houlsby2019parameter}
Neil Houlsby, Andrei Giurgiu, Stanislaw Jastrzebski, Bruna Morrone, Quentin De~Laroussilhe, Andrea Gesmundo, Mona Attariyan, and Sylvain Gelly.
\newblock Parameter-efficient transfer learning for nlp.
\newblock In \emph{International Conference on Machine Learning}, pages 2790--2799. PMLR, 2019.

\bibitem[Dettmers et~al.(2023)Dettmers, Pagnoni, Holtzman, and Zettlemoyer]{dettmers2023qlora}
Tim Dettmers, Artidoro Pagnoni, Ari Holtzman, and Luke Zettlemoyer.
\newblock Qlora: Efficient finetuning of quantized llms.
\newblock \emph{arXiv preprint arXiv:2305.14314}, 2023.

\bibitem[Hu et~al.(2021)Hu, Shen, Wallis, Allen-Zhu, Li, Wang, Wang, and Chen]{hu2021lora}
Edward~J. Hu, Yelong Shen, Phillip Wallis, Zeyuan Allen-Zhu, Yuanzhi Li, Shean Wang, Lu~Wang, and Weizhu Chen.
\newblock Lora: Low-rank adaptation of large language models, 2021.

\bibitem[Jain et~al.(2023)Jain, Chiang, Wen, Kirchenbauer, Chu, Somepalli, Bartoldson, Kailkhura, Schwarzschild, Saha, et~al.]{jain2023neftune}
Neel Jain, Ping-yeh Chiang, Yuxin Wen, John Kirchenbauer, Hong-Min Chu, Gowthami Somepalli, Brian~R Bartoldson, Bhavya Kailkhura, Avi Schwarzschild, Aniruddha Saha, et~al.
\newblock Neftune: Noisy embeddings improve instruction finetuning.
\newblock \emph{arXiv preprint arXiv:2310.05914}, 2023.

\bibitem[Susnjak et~al.(2022)Susnjak, Ramaswami, and Mathrani]{susnjak2022learning}
T.~Susnjak, G.S. Ramaswami, and A.~Mathrani.
\newblock Learning analytics dashboard: a tool for providing actionable insights to learners.
\newblock \emph{International Journal of Educational Technology in Higher Education}, 19\penalty0 (12), 2022.
\newblock \doi{10.1186/s41239-021-00313-7}.

\bibitem[Jiang et~al.(2023)Jiang, Sablayrolles, Mensch, Bamford, Chaplot, de~las Casas, Bressand, Lengyel, Lample, Saulnier, Lavaud, Lachaux, Stock, Scao, Lavril, Wang, Lacroix, and Sayed]{jiang2023mistral}
Albert~Q. Jiang, Alexandre Sablayrolles, Arthur Mensch, Chris Bamford, Devendra~Singh Chaplot, Diego de~las Casas, Florian Bressand, Gianna Lengyel, Guillaume Lample, Lucile Saulnier, Lélio~Renard Lavaud, Marie-Anne Lachaux, Pierre Stock, Teven~Le Scao, Thibaut Lavril, Thomas Wang, Timothée Lacroix, and William~El Sayed.
\newblock Mistral 7b, 2023.

\bibitem[Thorne et~al.(2018)Thorne, Vlachos, Christodoulopoulos, and Mittal]{thorne2018FEVER}
James Thorne, Andreas Vlachos, Christos Christodoulopoulos, and Arpit Mittal.
\newblock Fever: a large-scale dataset for fact extraction and verification.
\newblock \emph{ArXiv}, abs/1803.05355, 2018.
\newblock \doi{10.18653/v1/N18-1074}.

\bibitem[Yao et~al.(2022)Yao, Zhao, Yu, Du, Shafran, Narasimhan, and Cao]{yao2022ReAct}
Shunyu Yao, Jeffrey Zhao, Dian Yu, Nan Du, I.~Shafran, Karthik Narasimhan, and Yuan Cao.
\newblock React: Synergizing reasoning and acting in language models.
\newblock \emph{ArXiv}, abs/2210.03629, 2022.

\bibitem[Bekoulis et~al.(2020)Bekoulis, Papagiannopoulou, and Deligiannis]{bekoulis2020A}
Giannis Bekoulis, C.~Papagiannopoulou, and N.~Deligiannis.
\newblock A review on fact extraction and verification.
\newblock \emph{ACM Computing Surveys (CSUR)}, 55:\penalty0 1 -- 35, 2020.
\newblock \doi{10.1145/3485127}.

\bibitem[McIntosh et~al.(2024)McIntosh, Susnjak, Liu, Watters, and Halgamuge]{mcintosh2024inadequacy}
Timothy~R McIntosh, Teo Susnjak, Tong Liu, Paul Watters, and Malka~N Halgamuge.
\newblock The inadequacy of reinforcement learning from human feedback-radicalizing large language models via semantic vulnerabilities.
\newblock \emph{IEEE Transactions on Cognitive and Developmental Systems}, 2024.

\bibitem[McIntosh et~al.(2023{\natexlab{b}})McIntosh, Liu, Susnjak, Alavizadeh, Ng, Nowrozy, and Watters]{mcintosh2023harnessing}
Timothy McIntosh, Tong Liu, Teo Susnjak, Hooman Alavizadeh, Alex Ng, Raza Nowrozy, and Paul Watters.
\newblock Harnessing gpt-4 for generation of cybersecurity grc policies: A focus on ransomware attack mitigation.
\newblock \emph{Computers \& security}, 134:\penalty0 103424, 2023{\natexlab{b}}.

\end{thebibliography}

\appendix

\clearpage

\section{Qualitative Analysis Reference Q\&A Examples}
\label{qualitativeresponses}

\begin{figure}[hbt]
    \centering
    \includegraphics[width=1.0\textwidth]{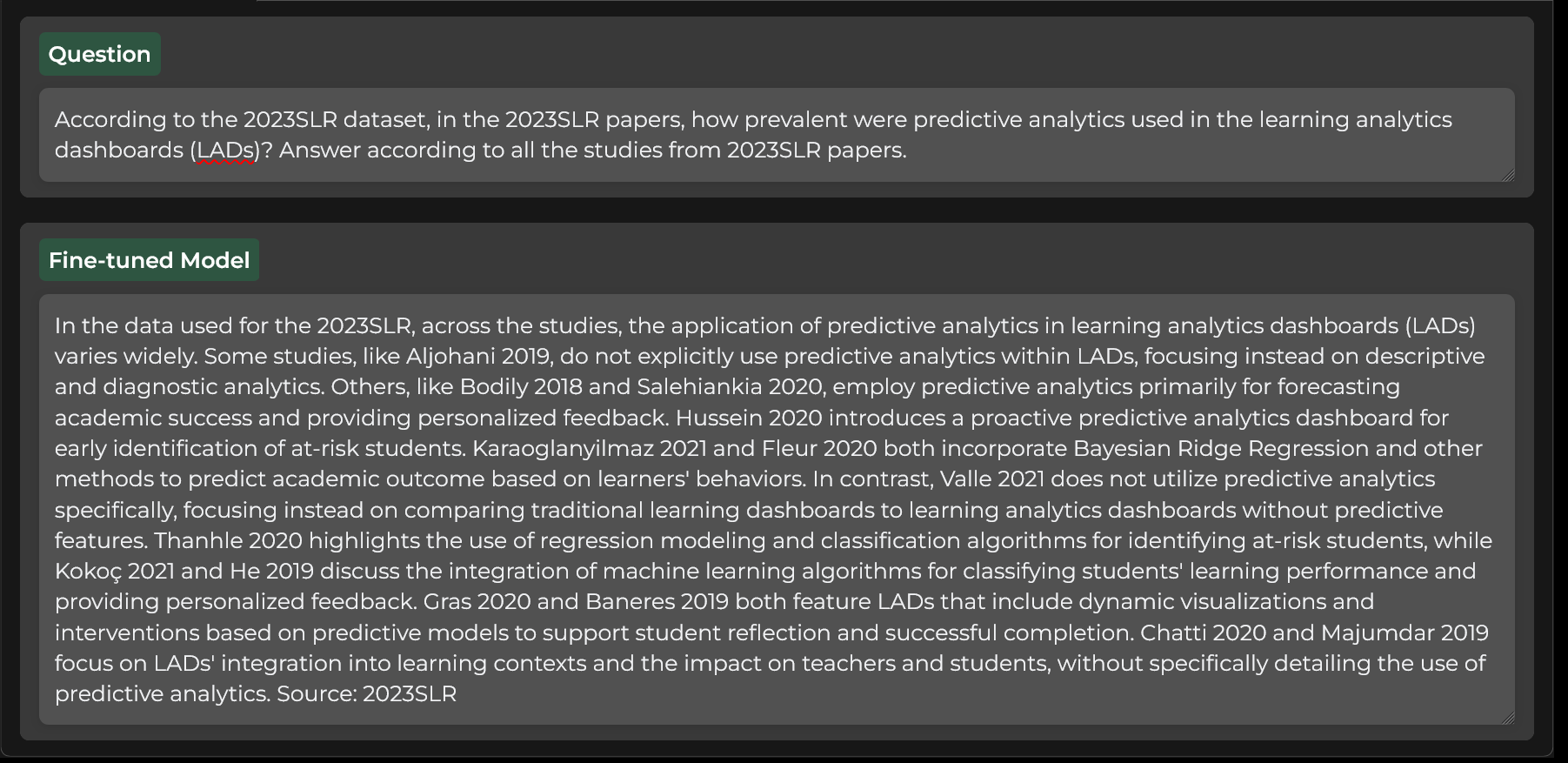}
    \caption{Example of a correct response. The  reference answer from the SLR study is "Predictive modelling has not featured in a larger percentage of reviewed LAD studies."}
    \label{fig:qa_correct1}
\end{figure}

\begin{figure}[hbt]
    \centering
    \includegraphics[width=1.0\textwidth]{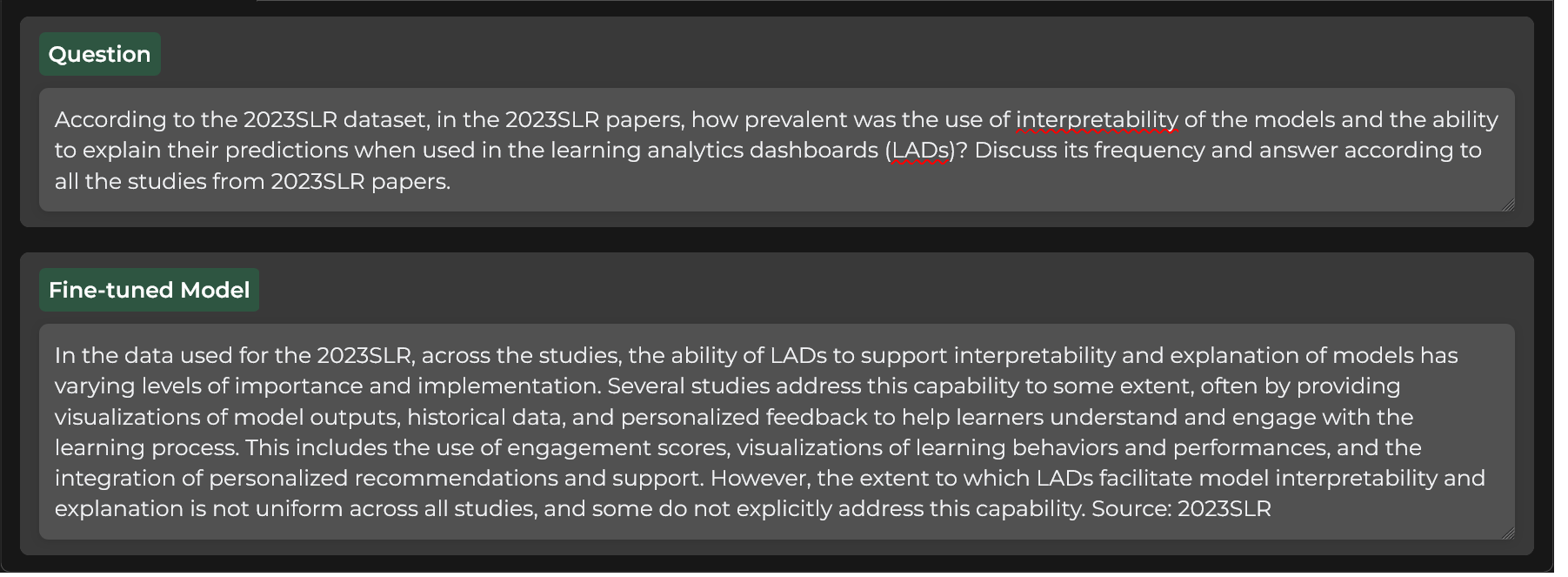}
    \caption{Example of a correct response. The reference answer from the SLR study is "We find that predictive modeling functionalities are not used in majority of cases within the reviewed LADs, and examples of interpretability of the models and the ability to explain their predictions to the learners do not yet exist in published studies."}
    \label{fig:qa_correct2}
\end{figure}

\begin{figure}[hbt]
    \centering
    \includegraphics[width=1.0\textwidth]{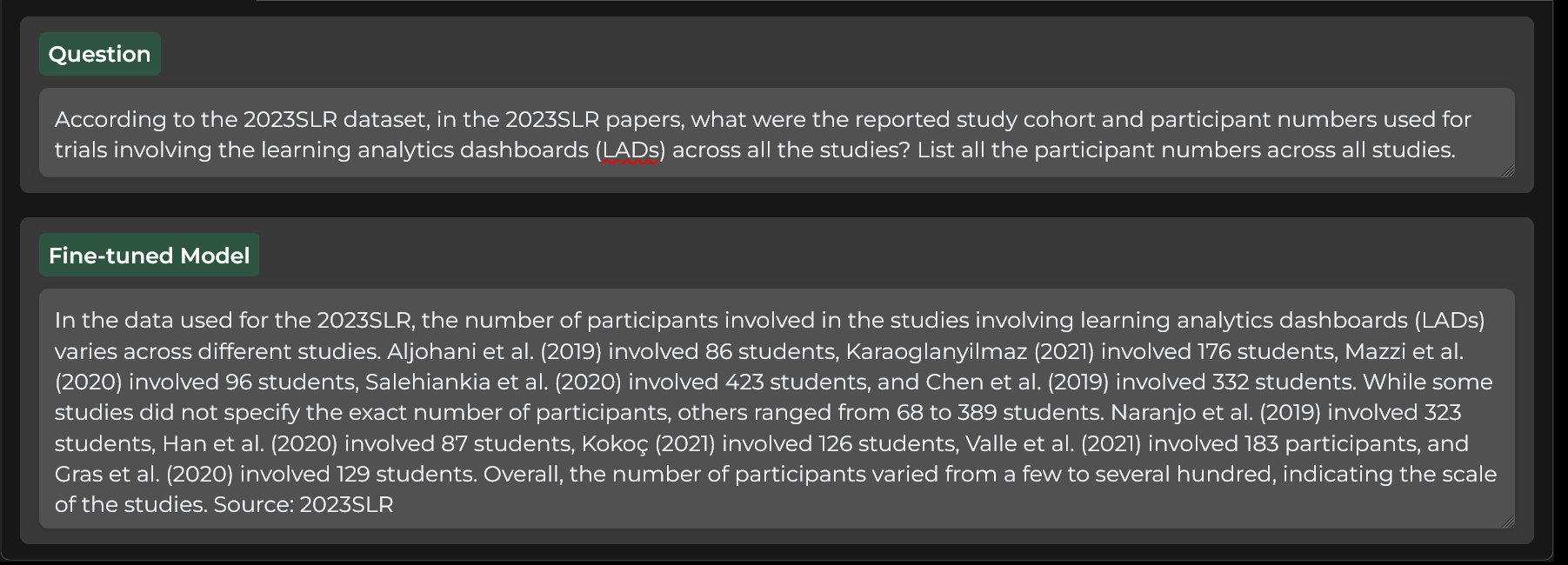}
    \caption{Example of an incorrect response. The reference answer from the SLR study is  "Bodily et al., 2018 (180), Chen et al., 2019, Aljohani et al., 2019 (86), Ulfa et al., 2019 (67), Majumdar et al., 2019, He et al., 2019 (327), Naranjo et al., 2019 (64), Baneres et al., 2019 (247), Gras et al., 2020 (127), Karaoglan Yilmaz \& Yilmaz, 2020 (81), Fleur et al., 020 (79), Chatti et al., 2020 (414), Kia et al., 2020 (449), Owatari et al., 2020 (108), Han et al., 2021 (88), Kokoç \& Altun, 2021 (126), Valle et al., 2021 (179)"}
    \label{fig:qa_incorrect}
\end{figure}

\section{LLM Evaluation Prompts}
\label{evaluationprompts}
This section illustrates the GPT-4 prompts used to evaluate all responses according to the FEVER and CGS metrics.
\clearpage

FEVER evaluation prompt:
\begin{mdframed}[backgroundcolor=gray!10]
\begin{verbatim}
Your task is to rigorously evaluate the factual accuracy, completeness 
and the source information of a given response text against a provided 
ground truth reference text using the FEVER categories from the Natural 
Language Processing field. This evaluation is critical for ensuring the 
reliability and quality of information synthesis systems. Please follow 
these detailed guidelines precisely:

Carefully read and understand the ground truth reference text in its entirety. 
Identify all key facts, details, findings, conclusions, and other important 
information presented.
Thoroughly read the response text multiple times if needed. Break it down 
into distinct claims, statements, or pieces of information.
For each distinct claim or piece of information in the response text, rigorously 
assess its factual status against the ground truth reference:

SUPPORTED: The claim and information in the reponse text are clearly, directly, 
completely and comprehensively corroborated and supported by evidence in the 
reference text. Also, the 'Source' information describing where the response 
originated from is identical.

NOT ENOUGH INFO: There is no evidence in the reference text that can either 
support or contradict the claims in the response text. The response text lacks 
enough information to make a judgement about consistency or contradictions with 
respect to the reference text.

REFUTED: Either the claims in the response text are completely, unambiguously 
and blatantly contradicted or refuted by evidence in the reference text, and 
he details in the response text are inaccurate with respect to the reference 
text. Or, the claims are supported between the texts, but the Source information 
describing where the response originated from is NOT THE SAME between the two texts.

Evaluate the degree of consistency as well as agreement between the following 
texts, and respond with a single FEVER categorical label as defined above. Be 
utterly objective, impartial and rigorous in your evaluation. Base your 
assessments solely on the information given, without making unsubstantiated 
assumptions or inferences. Check carefully the consistency between the Source 
information of the texts.

Reference Text: "<insert text >"

Response Text: "<insert text >"

Respond below with just a single FEVER categorical label.

Evaluation (SUPPORTED or REFUTED or NOT ENOUGH INFO):

\end{verbatim}\label{fever_eval_promt}
\end{mdframed}

\newpage
Consistency Grading Scale (CGS) quantitative evaluation prompt:

\begin{mdframed}[backgroundcolor=gray!10]
\begin{verbatim}

Your task is to rigorously evaluate the factual accuracy, completeness and the source
information of a given response text against a provided ground truth reference text. 
This evaluation is critical for ensuring the reliability and quality of information 
synthesis systems. Please follow these detailed guidelines precisely:

Carefully read and understand the ground truth reference text in its entirety. Identify 
all key facts, details, findings, conclusions, and other important information presented.
Thoroughly read the response text multiple times if needed. Break it down into distinct 
claims, statements, or pieces of information.
For each distinct claim or piece of information in the response text, rigorously assess 
its factual status against the ground truth reference:

FULLY CONSISTENT (2): The claim and information in the reponse text are clearly, 
directly,  completely and comprehensively corroborated by evidence in the reference text. 
Also, the  'Source' information describing where the response originated from is 
identical.

PARTIALLY CONSISTENT (1): Parts of the response text are supported by the reference text, 
but some details are either somewhat inaccurate, missing, too general or not fully 
substantiated from the reference text. Only the Source information describing where the 
response originated from is inconsistent.

NOT ENOUGH INFO (0): There is no evidence in the reference that can either support or 
contradict the claims in the response text. The response text lacks enough information 
to make a judgement about consistency or contradictions with respect to the reference 
text.

PARTIALLY CONTRADICTORY (-1): Parts of the response text directly contradict or are 
inconsistent with evidence present in the reference text, suggesting factual 
inaccuracies.

CONTRADICTORY (-2): The claims in the response text are completely, unambiguously and 
blatantly contradicted or refuted by evidence in the reference text. The details in the 
response text are inaccurate with respect to the reference text.

Evaluate the degree of consistency between the following texts, including the Source 
information of the texts, and respond with a single numerical value ranging from -2 to 
2 as defined above.  
Be utterly objective, impartial and rigorous in your evaluation. Base your assessments 
solely on the information given, without making unsubstantiated assumptions or 
inferences.

Reference Text: "<insert text >"

Response Text: "<insert text >"

Respond below with just a single number.

Evaluation (-2,-1,0,1,2):

\end{verbatim}\label{q_eval_promt}
\end{mdframed}

\end{document}